\documentclass[letterpaper]{article} 
\usepackage{aaai2026}  
\usepackage{times}  
\usepackage{helvet}  
\usepackage{courier}  
\usepackage[hyphens]{url}  
\usepackage{graphicx} 
\usepackage{natbib}  
\usepackage{caption} 
\usepackage{algorithm}
\usepackage{algorithmic}

\usepackage{booktabs}
\usepackage{amsmath}
\usepackage{subfig}
\usepackage{multicol}
\usepackage{multirow}
\usepackage{bm}
\usepackage{newfloat}
\usepackage{listings}
\DeclareCaptionStyle{ruled}{labelfont=normalfont,labelsep=colon,strut=off} 
\floatstyle{ruled}
\newfloat{listing}{tb}{lst}{}
\floatname{listing}{Listing}
\title{Enhanced Structured Lasso Pruning with Class-wise Information}
\author{
Xiang Liu\textsuperscript{1,*},
Mingchen Li\textsuperscript{2,*},
Xia Li\textsuperscript{3,$\dagger$},
Leigang Qu\textsuperscript{1},
Guansu Wang\textsuperscript{4},
Zifan Peng\textsuperscript{5},\\
Yijun Song\textsuperscript{6},
Zemin Liu\textsuperscript{7},
Linshan Jiang\textsuperscript{1,$\dagger$},
Jialin Li\textsuperscript{1}
}

\affiliations{
\textsuperscript{1}National University of Singapore,
\textsuperscript{2}University of North Texas\\
\textsuperscript{3}Department of Information Technology and Electrical Engineering, ETH Zürich\\
\textsuperscript{4}University of Melbourne,
\textsuperscript{5}Hong Kong University of Science and Technology (Guangzhou)\\
\textsuperscript{6}Dongfang College, Zhejiang University of Finance \& Economics\\
\textsuperscript{7}College of Computer Science and Technology, Zhejiang University\\
* Equal Contribution \quad
$\dagger$ Corresponding Authors\\[1mm]
\{liuxiang,lijl\}@comp.nus.edu.sg, mingchenli@my.unt.edu, xiaxiali@ethz.ch, guansuw@student.unimelb.edu.au\\
leigangqu@gmail.com, zpengao@connect.hkust-gz.edu.cn, yijunsong.0377@gmail.com,\\
liu.zemin@hotmail.com, linshan@nus.edu.sg
}

\usepackage{bibentry}

\begin{document}

\maketitle

\begin{abstract}
Modern applications require lightweight neural network models. Most existing neural network pruning methods focus on removing unimportant filters; however, these may result in the loss of statistical information after pruning due to failing to consider the class-wise information. In this paper, we employ the structured lasso from the perspective of utilizing precise class-wise information for model pruning with the help of Information Bottleneck theory, which guides us to ensure the retention of statistical information before and after pruning. With these techniques, we propose two novel adaptive network pruning schemes in parallel: sparse graph-structured lasso pruning with Information Bottleneck (\textbf{sGLP-IB}) and sparse tree-guided lasso pruning with Information Bottleneck (\textbf{sTLP-IB}). The key component is that we prune the model filters utilizing sGLP-IB and sTLP-IB with more precise structured class-wise relatedness. Compared to multiple state-of-the-art methods, our approaches achieve the best performance across three datasets and six model structures on extensive experiments. For example, with the VGG16 model based on the CIFAR-10 dataset, we can reduce the parameters by 85\%, decrease the FLOPs by 61\%, and maintain an accuracy of 94.10\% (0.14\% better than the original). For large-scale ImageNet, we can reduce the parameters by 55\% while keeping the accuracy at 76.12\% (only drop 0.03\%) using the ResNet architecture. In summary, we succeed in reducing the model size and computational resource usage while maintaining the effectiveness of accuracy. 
\end{abstract}


\section{Introduction}

Deep Learning~\cite{Krizhevsky2012ImageNetCW}, especially CNN-based methods~\cite{Simonyan2014VeryDC,Szegedy2014GoingDW,He2015DeepRL}, has proven its effectiveness in tasks in numerous applications, such as multimodal fusion and vision. Recently, researchers have been attempting to deploy these neural network models on resource-constrained devices, like smartphones, for users' real-time analysis of multi-modal information~\cite{Liu2024EDViTSV,ijcai2024p596}. However, the number of these network parameters is often too large, making the deployment difficult. For instance, the VGGNet-16 model has parameters exceeding 500MB~\cite{Simonyan2014VeryDC}, which introduces a heavy memory burden and computation overhead. Current researchers aim to explore methods that help deploy models on resource-constrained devices.

Fortunately, scientists have discovered that the parameters are usually redundant and the key weights of neural networks exhibit sparsity characteristics~\cite{He2017ChannelPF}. As a result, a significant amount of work has focused on pruning unimportant channels to reduce the size of the model, which reduces the model's FLOPs and lowers computational overhead as well. The common methods can be primarily divided into two main categories: unstructured pruning and structured pruning~\cite{Shao2022StructuredPF}. Unstructured pruning mainly focuses on weight pruning~\cite{Ding2019GlobalSM} and N:M sparsity~\cite{Zhang2022LearningBC}. However, both of these methods require hardware support, which makes them less scalable. Therefore, this paper primarily focuses on structured pruning, which benefits from removing unimportant filters within the CNNs and is consistent with the focus of most current pruning research.

To avoid the limitations of hand-crafted approaches, most structured pruning methods utilize statistical information~\cite{Zhang2021ExplorationAE,Shao2022StructuredPF} to analyze filters insides CNNs for \textbf{channel selection}, with the aim of minimizing the loss before and after pruning and achieving better accuracy results. However, while most existing methods overlook class-wise information~\cite{Hu2016NetworkTA}, some techniques do incorporate inter-class information~\cite{hemmat2020cap,hu2023catro,ma2023ocap}; however, they fail to model the finer-grained class relatedness, resulting in a less precise pruning scheme and a consequent drop in accuracy. Furthermore, an ideal pruning method should also include an adaptive approach~\cite{guo2023automatic,Xie2024AdaptivePO} to control the \textbf{sparsity ratio} according to specific requirements without the need to manually adjust hyperparameters.

Based on these, we thus employ the precise Information Bottleneck (IB)~\cite{Tishby2000TheIB} theory perspective to model statistical information of feature maps of CNN for channel selection, aiming to minimize the loss from pruning. We also utilize structured lasso techniques that account for sparsity, such as graph-structured lasso~\cite{Liu2024NovelTG} and tree-guided lasso~\cite{Kim2009TreeGuidedGL,Liu2024SparseVS}, which can also consider the relatedness in feature maps and form corresponding sub-groups and sub-trees. This approach helps to incorporate more class-wise information compared to previous methods. Consequently, we propose two novel methods: Sparse Graph-structured Lasso Pruning with Information Bottleneck (\textbf{sGLP-IB}) and Sparse Tree-guided Lasso Pruning with Information Bottleneck (\textbf{sTLP-IB}). Our methods perform pruning at the layer level, ensuring more precise pruning filters and preventing cumulative errors layer by layer. Based on this, our achievements have already been surprising as we are the first to integrate the precise structured class-wise information into pruning. Further, we  (1) combine Gram matrix~\cite{Lanckriet2002LearningTK} to reduce the dimensions and accelerate computation without compromising accuracy, (2) propose an adaptive method to ensure that our method can be automatically suitable for varying sparsity ratio requirements rather than setting hyperparameters manually based on the expert experiences, (3) use novel optimization method which has polynomial time complexity with guaranteed convergence rate. We conduct extensive experiments to evaluate the effectiveness.

Our main contributions are summarized as follows: 

\begin{itemize}
    \item We are the first to discover pruning with structured lasso from the information bottleneck perspective via precise structured class-wise information. By enhancing techniques such as graph-structured lasso and tree-guided lasso with information bottleneck, we propose two novel methods for model pruning: sGLP-IB and sTLP-IB.
    
    \item We have optimized our proposed methods, sGLP-IB and sTLP-IB, to ensure faster computation using Gram matrix and adaptively address different model sizes or FLOPs requirements with our proposed adaptive algorithm. These optimizations still guarantee convergence and polynomial computation time complexity using the proximal gradient descent method.
    
    \item We conduct comparative experiments against numerous baselines using three widely used datasets, CIFAR-10, CIFAR-100 and ImageNet. We tested multiple variants of VGGNet, ResNet, and GoogleNet. The experimental results demonstrate the effectiveness of our methods across three metrics: accuracy, memory size, and FLOPs. They both achieve state-of-the-art results, while sTLP-IB demonstrates a stronger ability to learn class information compared to sGLP-IB. Extensive ablation studies further confirm the robustness of our methods and the potential of utilizing structured class-wise information for model pruning.
\end{itemize}


\section{Related Works}
\label{sec:related}

\subsection{Structured Pruning}
\label{sec:structured}
MPF~\cite{He2019FilterPB} considers the geometric distance between filters and neighboring filters to guide the pruning process. CPMC~\cite{Yan2020ChannelPV} leverages relationships between filters in the current layer and subsequent layers for pruning decisions. HRank~\cite{Lin2020HRankFP} compresses models by constructing low-rank matrices using information from the same batch. CPGMI~\cite{Lee2020ChannelPV} leverages gradients of mutual information to measure the importance of channels to prune the less important ones. SCOP~\cite{Tang2020SCOPSC} reduces filters based on their deviation from the expected network behavior. 
GDP~\cite{Guo2021GDPSN} employs gates with differentiable polarization, learning whether gates are zero during training for pruning purposes. EEMC~\cite{Zhang2021ExplorationAE} uses a multivariate Bernoulli distribution along with stochastic gradient Hamiltonian Monte Carlo for pruning. SRR-GR~\cite{Wang2021ConvolutionalNN} identifies redundant structures within CNNs. FTWT~\cite{Elkerdawy2021FireTW} predicts pruning strategies using a self-supervised mask. DECORE~\cite{Alwani2021DECOREDC} assigns an agent to each filter and uses lightweight learning to decide whether to keep or discard each filter. EPruner~\cite{Lin2021NetworkPU} employs Affinity Propagation for efficient pruning. DPFPS~\cite{Ruan2021DPFPSDA} directly learns the network with structural sparsity for pruning. CC~\cite{Li2021TowardsCC} combines tensor decomposition for filter pruning. NPPM~\cite{Gao2021NetworkPV} trains the network to predict the performance of the pruned model, guiding the pruning process. LRF-60~\cite{Joo2021LinearlyRF} uses Linearly Replaceable Filters to aid pruning decisions. AutoBot~\cite{Castells2021AutomaticNN} tests each filter one-by-one to ensure pruning precision. 
PGMPF~\cite{Cai2022PriorGM} utilizes prior masks as the basis for pruning masks. DLRFC~\cite{He2022FilterPV} uses Receptive Field Criterion to measure filter importance. FSM~\cite{Duan2022NetworkPV} aggregates feature and filter information to evaluate their relevance for shifting. Random~\cite{Li2022RevisitingRC} employs random search strategies for pruning.  The study~\cite{Hussien2024SmallCS} employs activation statistics, foundations in information theory, and statistical analysis with designed regularizations for pruning. CSD~\cite{Xie2024AdaptivePO} uses channel spatial dependability metric and leverages feature characteristics for pruning. However, the statistical learning information considered by these methods may not be precise enough.

\subsection{Information Bottleneck and Lasso Principle}

The Information Bottleneck (IB)~\cite{Tishby2000TheIB} method extracts the most relevant output information by considering the input. Aside from the previous mentioned methods, the works~\cite{Tishby2015DeepLA,Dai2018CompressingNN} are the first to use IB for compressing CNNs. FPGM~\cite{He2018FilterPV} employs the geometric median for pruning. Li et al.~\cite{Li2016PruningFF} illustrate the correspondence between Lasso and IB for pruning. The approach in~\cite{He2017ChannelPF} utilizes a two-step optimization process incorporating Lasso for pruning. Papers~\cite{Scardapane2016GroupSR,Rajaraman2023GreedyPW} leverage the $l_2$ lasso (e.g., group lasso) for pruning, with analyzed complexity~\cite{Rajaraman2023GreedyPW}. APIB~\cite{guo2023automatic} integrates IB with the Hilbert-Schmidt Independence Criterion Lasso for exploration. The paper~\cite{Sakamoto2024EndtoEndTI} also shows through analysis that layer-level pruning, based on the Hilbert-Schmidt independence criterion, is preferable to end-to-end pruning.

\textbf{Motivation: }However, none of these methods account for class-wise information. To address this, from the information bottleneck perspective, we employ state-of-the-art techniques like graph-structured lasso~\cite{Liu2024NovelTG} and tree-guided lasso~\cite{Liu2024SparseVS} to aggregate structured class-wise information effectively to propose our novel methods and prune filters inside  CNNs. The following section explains how we combine Information Bottleneck with structured Lasso as well as reducing computational complexity without compromising accuracy, and how we integrate our adaptive method with our novel methods to control the sparsity.

\begin{figure*}[!th]
  \centering
  \includegraphics[width=0.88\textwidth]{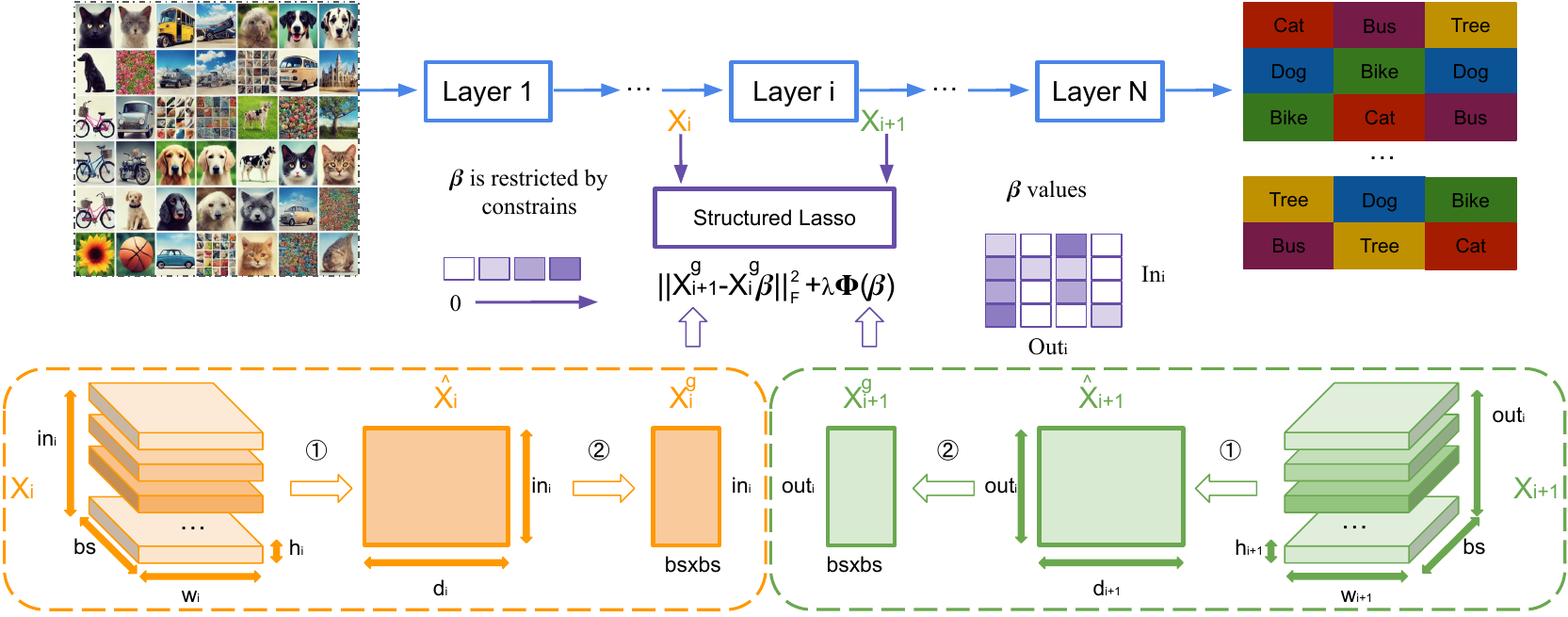}
  \vspace{-0.5em}
  \caption{The overview of our methods. To utilize information bottleneck theory, \textcircled{1} reshape the input and output feature maps. \textcircled{2} transforms the reshaped feature maps with Gram matrix. Then, we utilize the structured lasso to account for the structured class-wise information to get the linkage matrix and prune the filters.}
  \label{fig:overview}
    \vspace{-1.3em}
\end{figure*}

\section{Methodology}
\label{sec:method}
\subsection{Design Overview}

The overview of our methods is shown in Fig.~\ref{fig:overview}. To avoid the imprecision that end-to-end methods might introduce, we perform layer-wise pruning independently. We reshape each layer's input and output feature map tensors into two-dimensional matrices, enabling us to apply the information bottleneck theory to determine which filters can be pruned. To further reduce computational complexity, capture internal geometric structures, and utilize more flexible kernel functions, we employ Gram matrix~\cite{Lanckriet2002LearningTK}. Since lasso aligns well with the information bottleneck theory~\cite{Tishby2015DeepLA,Dai2018CompressingNN}, we use structured lasso for solving this problem. This approach also takes into account structured class-wise information in the feature maps (as well as filters), resulting in a sparse linkage map between layers' feature maps for pruning.

\noindent \textbf{Math Notations:} In this paper, given a CNN model $M$ with $N$ layers, we let $\bm{C}_i$ denote the $i$-th layer of $M$. The input feature map of $\bm{C}_i$ is $\bm{X}_i$ with the shape $(bs,h_i,w_i,in_i)$; the output feature map of $\bm{C}_i$ is $\bm{X}_{i+1}$ with the shape $(bs,h_{i+1},w_{i+1},out_i)$, where $i \in \{1,2,...,N\}$, $\bm{X}_1$ is the input sample batch from the datasets, $\bm{X}_{N+1}$ is the final output results, $\bm{X}_{i+1}$ is the input of $\bm{C}_{i+1}$; for layer $i$, $bs$ is the batch size, $h_i$ and $w_i$ are the height and width of each feature map, $in_i$ is the input channel, $out_i$ is the output channel (filter number of $\bm{C}_i$). We convert $\bm{X}_i$ from the original shape to $\hat{\bm{X}_i}$, $(bs\times h_i\times w_i, in_i)$, i.e., $(d_i,in_i)$. Thus, the shape of $\hat{\bm{X}_{i+1}}$ is $(bs\times h_{i+1}\times w_{i+1}, out_i)$, i.e., $(d_{i+1},out_i)$. Therefore, we use $\bm{\beta}$ to denote the linkage matrix of $\bm{C}_i$ among $\bm{X}_i$ and $\bm{X}_{i+1}$, $\bm{\beta}_j$ denotes the $j$-th column $\bm{\beta}$.

\subsection{Structured Lasso}

\subsubsection{Information Bottleneck Function and Gram matrix}
For each layer $\bm{C}_i$, we optimize on the Information Bottleneck (IB) formular~\cite{Tishby2000TheIB} as:
{\small
\begin{equation}
\label{equa:IBbasic}
    \min \limits_{\Tilde{\bm{X}_{i+1}}}-I(\Tilde{\bm{X}_{i+1}};\bm{X}_{i})+\gamma I(\bm{X}_{i+1};\Tilde{\bm{X}_{i+1}})
\end{equation}}
where $\Tilde{\bm{X}_{i+1}}$ denotes the pruned ouput features and $\gamma$ is a positive Largrange multiplier. Output feature maps are derived from input feature maps through filters and corresponding convolution operations. As such, important filters result in output feature maps that contain more information from the corresponding input feature maps. To better utilize the IB theory, we first reshaped $\bm{X}_i$ and $\bm{X}_{i+1}$ into two-dimensional matrices, denoted as $\hat{\bm{X}_i}$ and $\hat{\bm{X}_{i+1}}$. Therefore, the problem of pruning filters is transformed into the task of exploring to preserve output feature channels that contain the most statistical information with respect to the input feature maps using the Information Bottleneck approach. 
{\small
\begin{equation}
\label{equa:IBreshape}
    \min \limits_{\Tilde{\bm{X}_{i+1}}}-I(\Tilde{\bm{X}_{i+1}};\hat{\bm{X}_{i}})+\gamma I(\hat{\bm{X}_{i+1}};\Tilde{\bm{X}_{i+1}})
\end{equation}}
Note that Gram matrix allows for the use of more flexible kernel functions without the loss of information, enabling linear operations in high-dimensional spaces, thereby reducing computational complexity and better considering the geometric structure between feature maps. Thus, to better account for structured class-wise information between feature maps and reduce computation complexity, we employ Gram matrix. We transform $\hat{\bm{X}_i}$ and $\hat{\bm{X}_{i+1}}$ into centered Gram matrices $\bm{X}_{i}^g$ and  $\bm{X}_{i+1}^g$. Therefore, we have the optimization formula:
{\small
\begin{equation}
\label{equa:IBgram}
    \min \limits_{\Tilde{\bm{X}_{i+1}}}-I(\Tilde{\bm{X}_{i+1}};\bm{X}_{i}^g)+\gamma I(\bm{X}_{i+1}^g;\Tilde{\bm{X}_{i+1}})
\end{equation}}
where $\bm{X}_i^g=\bm{\Psi} \bm{X}_i^k \bm{\Psi}$, $\bm{\Psi}=I_{bs}-\frac{1}{bs}1_{bs}1_{bs}^T$, $I_{bs}$ is a bs-dimensional identity matrix, $1_{bs}$ is a  bs-dimensional vector consisting entirely of ones. The element in row $a$ and column $b$ of  $\bm{X}_i^k$ is equal to the inner product of $a$-th feature map and $b$-th feature map of $\bm{X}_i$. Therefore, we have $\bm{X}_i^g \in R^{bs\times bs, in_i},\bm{X}_{i+1}^g \in \bm{R}^{bs\times bs, out_i}, \bm{\beta} \in \bm{R} ^{in_i, out_i}$. 

The first step of reshaping is intended to capture the relationship between input features and output features from the channel perspective. The second step is to utilize Gram matrix. In practice, we can directly transform $\bm{X}_i$ to $\bm{X}_i^g$.

\subsubsection{Graph-structured Lasso and sGLP-IB}
The computation of the IB method is often considered computationally infeasible due to the significant overhead introduced by density estimation in high-dimensional spaces~\cite{Ma2019TheHB}. We are able to use the Lasso-based methods to get the nonlinear dependencies between the input and output of each layer~\cite{Li2016PruningFF,guo2023automatic}.
{\small
\begin{equation}
\label{equa:standardlasso}
\bm{\beta} = \min  \limits_{\bm{\beta}}\frac{1}{2}||\bm{X}_{i+1}^g - \bm{X}_i^g\bm{\beta}||_F^2 + \Phi(\bm{\beta})
\end{equation}}
where $||\cdot||_F$ denotes the Frobenius norm, and $\Phi$ is determined by structured lasso.
Due to the fact that filters generally contain class-wise information~\cite{Hu2016NetworkTA}. Filters are tailored to different classes, meaning they also possess distinct statistical properties. As a result, it is possible to form corresponding subgroups based on class information. Thus, the output feature maps also conceal structured class-wise information due to the filters. This results in a complex relationship between output feature maps, especially in terms of statistical subgroups. Therefore, we introduce graph-structured lasso from structured lasso to better model this relationship.

Given $\bm{\beta}$ for $\bm{C}_i$, we have the node sets $V$, \{$\bm{\beta}_1, \bm{\beta}_2, ...\bm{\beta}_{out_i}$\} that forms the graph, where $\bm{\beta}_j^r$ denotes element in $j$-th column and $r$-th row of $\bm{\beta}$. We have weight edge set $E$ of the graph by computing pairwise Pearson correlations and linking pair of nodes when their correlation exceeds a threshold $th$, where the weight of each edge indicates correlation degree. Let $f_{lm}$ represent the weight of edge $e = (l, m) \in E$, which quantifies the correlation between element $l$ and $m$.  With these, we define graph-structured lasso as:
{\small
\begin{equation}
\label{GFlasso}
\Phi(\bm{\beta}) = \lambda||\bm{\beta}||_1 + \mu\sum_{e=(l,m) \in E} |f_{lm}|\sum_{j=1}^{in_i}|\bm{\beta}_{l}^j - sign(f_{lm})\bm{\beta}_{m}^j|
\end{equation}}
where $\lambda$ is the hyperparameter for the sparsity and $\mu$ regulates inner class-wise dependency. Substituting \eqref{GFlasso} into \eqref{equa:standardlasso} yields our sparse graph-structured lasso
pruning with Information Bottleneck (sGLP-IB) method.

As shown in the $\bm{\beta}$ matrix in Fig.~\ref{fig:overview}, each column of $\bm{\beta}$ represents the relatedness between all feature maps of this batch for a specific output channel and the current layer's input feature maps. Our method can consider output feature maps and $\bm{\beta}$, forming subgroups according to the classes. Within each subgroup, we identify and select the filters that contribute the most. The larger the value of elements in $\bm{\beta}$, the stronger the correlation. Therefore, we can determine whether to retain a filter based on the number of non-zero values in its corresponding column.

\subsubsection{Tree-Guided Lasso and sTLP-IB}
We also apply the state-of-the-art tree-guided lasso penalty to model the more closely related structured class-wise information and then form the tree structures. The tree numbers and the overlap among groups are determined by the hierarchical clustering tree due to class information.  We have:
{\small
\begin{equation}
\label{euqation: treelassopenalty}
\Phi(\bm{\beta}) = \lambda\sum_{j} W_j(v_{root})= \lambda\sum_{j} \sum_{v\in V} w_{ve}||\bm{\beta}_j^{G_v}||_2 
\end{equation}}
\noindent Where each group of $j$-th subtree regression factors $\bm{\beta}_j^{G_v}$ is weighted by $w_{ve}$, as in \eqref{w_v}, where $A$ is ancestor set. Vector $\bm{\beta}_j^{G_v}$ represents regression factors $\{\bm{\beta}_j^k \mid k \in G_v\}$.  Each node $v \in V$ of one tree is associated with a group $G_v$, whose members are columns of $\bm{\beta}$ within nodes of the same subtree.
{\small
\begin{equation}
\label{w_v}
{   
w_{ve}=\left\{
  \begin{array}{rcl}
    (1-h_v)\prod\limits_{q\in A(v)}{h_q}&\mbox{ $v \in$  internal node}\\
    \prod\limits_{q\in A(v)}{h_q}&\mbox{$v \in$ leaf node}
  \end{array}
\right. 
}
\end{equation}}
Generally,  $h_v$ represents the weight for selecting covariates relevant specifically to column of $\bm{\beta}$ linked to each child of node $v$, while $1 - h_v$ is the weight for jointly selecting covariates relevant to all children of node $v$, $h_v \in (0,1)$. Assuming $|V|$ represents the number of nodes in a tree, since a tree associated with $out_i$ columns of $\bm{\beta}$ for $\bm{C}_i$ has $2out_i - 1$ nodes, $|V|$ in tree-lasso penalty is upper-bounded by $2out_i$.

Substituting \eqref{euqation: treelassopenalty} into \eqref{equa:standardlasso} yields our sparse tree-guided lasso pruning with Information Bottleneck (sTLP-IB) method. Similarly, within each subtree formed by classes, we select the filters that could convey more information.

\subsection{Optimization with Proximal Gradient Descent}

With smoothing proximal gradient descent method~\cite{chen2012smoothing,Yuan2024SmoothingPG} we implemented, we could easily apply into \eqref{equa:standardlasso} and yield the $\bm{\beta}$ results for each layer, which also guarantee the convergence rate of O($\frac{1}{\epsilon}$) with polynomial time complexity, where $\epsilon$ is the differences between the yielded $\bm{\beta}$ and the optimal $\bm{\beta}^*$.  We then prune the model $M$ based on the $\bm{\beta}$ obtained from sGLP-IB and sTLP-IB.

\begin{algorithm}[!t]
    \caption{Adaptive Method for Pruning}
    \label{alg:adaptive}
    \begin{flushleft}
    \textbf{Input}: Training Dataset Batch, Pretrained Model $M$ with $N$ layers, compressed memory size/FLOPs range $\{E_l, E_u\}$ \\
\textbf{Parameter}: $\lambda_l$, $\lambda_u$\\
\textbf{Output}: Pruned Model $M^*$
\end{flushleft}
    \begin{algorithmic}[1]
        \FOR{$i$ in $N$}
            \STATE $\lambda_{lt}=\lambda_l, \lambda_{ut}=\lambda_u$, $\bm{C}_i \leftarrow M$.
            \STATE $flag=True$.
            \WHILE{$flag$}
                \STATE $\lambda=log(\frac{1}{2}(exp(\lambda_{lt})+exp(\lambda_{ut})))$.
                \STATE $\bm{C}_i^*$= sGLP-IB$(\bm{C}_i,\lambda)$ or sTLP-IB$(\bm{C}_i,\lambda)$.
                 \IF { $size(\bm{C}_i^*) \leq E_l$ } 
                    \STATE $\lambda_{ut}=\lambda$.
                \ELSIF{ $size(\bm{C}_i^*) \geq E_u$}
                    \STATE $\lambda_{lt}=\lambda$.
                \ELSE
                    \STATE $flag=False$.
                \ENDIF
            \ENDWHILE      
            \STATE $M^* \leftarrow \bm{C}_i*$.
        \ENDFOR
        \STATE \textbf{return} $M^*$.
    \end{algorithmic}
\end{algorithm}

\subsection{Adaptive Algorithm}
We illustrate how our approach utilizes sGLP-IB and sTLP-IB to prune networks. We observe that they include a hyperparameter $\lambda$, which greatly influences the sparsity of $\bm{\beta}$. If $\lambda$ is too small, each element of $\bm{\beta}$ is a non-zero value, making the pruning ineffective; if  $\lambda$ is too large, $\bm{\beta}$ may contain too few non-zero values, making the pruning impossible as they will prune all the filters in one layer. Given that models have many layers, setting $\lambda$ for each layer can be very time-consuming. Therefore, we propose an adaptive method to selectively prune based on memory size and FLOPs requirements as shown in Algorithm~\ref{alg:adaptive}.

In the adaptive method, we use a binary search approach. We set an upper and lower limit for the value of $\lambda$ and use exponential mapping to quickly find an appropriate $\lambda$ that fits the sparsity level of the structured Lasso. Given a range of the proportion of parameters to retain or a range of FLOPs to retain, our algorithm can adaptively find suitable $\lambda$ values for each layer to ensure that only a specific number of columns have non-zero values in $\bm{\beta}$ and meet the requirements. This method ensures a relatively balanced number of parameters across layers, preventing any layer from having too many or too few parameters. It also allows for the setting of different parameter numbers and FLOP retention ratios for each layer.

\begin{table}[!t]
\centering
\resizebox{\linewidth}{!}{
\begin{tabular}{lcccc}
\toprule
\textbf{Method}  &\textbf{Top-1(\%)}    &\textbf{FLOPs}$\downarrow$ &\textbf{Params} $\downarrow$  &  $\Delta$ \textbf{Acc(\%)}   \\
\midrule
\texttt{VGGNet-16} &\texttt{93.96} & \texttt{0(314.57M)}  &\texttt{0(14.98M)}  &\texttt{0}      \\
CPMC      &93.40 & 66\%         & -        &-0.56  \\ 
L1        &93.41 & 34\%         &64\%      &-0.55  \\
HRank     &93.42 & 54\%         &83\%      &-0.54  \\
FPGM      &93.50 & 36\%         & -        &-0.46  \\           
PGMPF     &93.60 & 66\%         & -        &-0.36  \\
EEMC      &93.63 & 56\%         & -        &-0.33  \\
CSD       &93.69 & 45\%         & 48\%     &-0.27  \\
\textbf{sGLP-IB}   &\textbf{93.98} & \textbf{65\%}   &\textbf{91\%} &+0.02    \\
GDP       &93.99 & 31\%         & -        &+0.03  \\
APIB      &94.00 & 66\%         & 78\%     &+0.04  \\
\textbf{sTLP-IB}   &\textbf{94.01} & \textbf{64\%}   &\textbf{91\%} &+0.05    \\
DECORE    &94.02 & 35\%         & 63\%     &+0.06  \\
NORTON    &94.08 & 60\%         & 83\%     &+0.12  \\
AutoBot   &94.10 & 54\%         & 50\%     &+0.14  \\
\textbf{sGLP-IB}   &\textbf{94.05} & \textbf{61\%}   &\textbf{85\%} &+0.09    \\
\textbf{sTLP-IB}   &\textbf{94.10} & \textbf{61\%}   &\textbf{85\%} &+0.14    \\
\midrule
\texttt{ResNet-56} &\texttt{93.28} &\texttt{0(127.62M)} &\texttt{0(0.853M)}  &\texttt{0}\\
DECORE	&90.85	&81\%	&85\%	&-2.43\\
FTWT	&92.63	&66\%	&-	&-0.65\\
FSM	    &92.76	&68\%	&68\%	&-0.52\\
Hrank	&93.17	&50\%	&42\%	&-0.11\\
DLRFC	&93.57	&53\%	&55\%	&+0.29\\
SRR-GR	&93.75	&54\%	&-	&+0.47\\
CSD	    &93.89	&45\%	&40\%	&+0.61\\
APIB	&93.92	&47\%	&51\%	&+0.64\\
AutoBot	&93.94	&50\%	&44\%	&+0.66\\
NORTON	&94.00	&42\%	&48\%	&+0.72\\
\textbf{sGLP-IB}	&\textbf{93.93}	&\textbf{53\%}	&\textbf{52\%}	&\textbf{+0.65}\\
\textbf{sTLP-IB}	&\textbf{93.96}	&\textbf{53\%}	&\textbf{52\%}	&\textbf{+0.68}\\
\midrule
\texttt{ResNet-110} & \texttt{93.50} & \texttt{0(256.04M)} & \texttt{0(1.73M)} &\texttt{0}\\ 
DECORE	&92.71	&77\%	&80\%	&-0.79\\
Hrank	&93.36	&58\%	&59\%	&-0.14\\
APIB	&93.37	&77\%	&82\%	&-0.13\\
MPF	    &93.38	&63\%	&-	&-0.12\\
DECORE	&93.50	&61\%	&64\%	&0\\
EPruner	&93.62	&65\%	&76\%	&+0.12\\
NORTON	&93.68	&66\%	&68\%	&+0.18\\
\textbf{sGLP-IB}	&\textbf{93.69}	&\textbf{64\%}	&\textbf{68\%}	&\textbf{+0.19}\\
\textbf{sTLP-IB}	&\textbf{93.80}	&\textbf{62\%}	&\textbf{68\%}	&\textbf{+0.30}\\
\midrule
\texttt{GoogleNet} & \texttt{95.05} & \texttt{0(1531.98M)} & \texttt{0(6.17M)} & \texttt{0}\\
Hrank	&94.53	&55\%	&55\%	&-0.52\\
FSM	    &94.72	&63\%	&56\%	&-0.33\\
EPruner	&94.90	&63\%	&66\%	&-0.15\\
\textbf{sGLP-IB}	&\textbf{94.75}	&\textbf{69\%}	&\textbf{63\%}	&\textbf{-0.30}\\
\textbf{sTLP-IB}	&\textbf{94.78}	&\textbf{69\%}	&\textbf{63\%}	&\textbf{-0.27}\\
\bottomrule
\end{tabular}
}
\vspace{-0.5em}
\caption{Pruning results of VGGNet-16, ResNet-56, ResNet-110 and GoogleNet on CIFAR-10 dataset. For VGGNet-16, we present results under different pruning ratios.}
\vspace{-1.5em}
\label{table:main}
\end{table}

\section{Experiments}
\label{sec:experiments}

To demonstrate the effectiveness of our sGLP-IB and sTLP-IB, we conduct extensive experiments based on many representative CNNs, including various variants of VGGNet, ResNet, and GoogleNet, using three widely used datasets: CIFAR-10, CIFAR-100, and ImageNet. We also conduct experiments on resource-constrained devices (Raspberry Pi 4) and perform numerous ablation studies to analyze the robustness and characteristics of our methods.

\textbf{Implementation Details: }All experiments are implemented using PyTorch and averaged over five trials with random seeds. We use the SGD optimizer with a decaying learning rate initialized to 1$e$-4, and we set the batch size to 128. For sGLP-IB, we choose a threshold of 0.61,8 as graph-structured lasso. After pruning, for fair comparison, we finetune VGGNet, ResNet and GoogleNet for 400 to 500 epochs. On ImageNet datasets, we train VGGNet and ResNet for 150 to 200 epochs. We choose the Laplacian kernel as our base kernel function for Gram matrix. We prune without the first four layers.

\textbf{Evaluation Metric:} We evaluate the pruned model based on its Top-1 test accuracy, the pruned ratio of parameters, the reduced FLOPs ratio with the original pretrained model.

\textbf{Baselines:} Our extensive baselines are primarily based on the aforementioned structured pruning methods in Section~\ref{sec:structured}. Besides, to provide a more comprehensive evaluation, we include NORTON~\cite{Pham2024EfficientTD,Pham2024EnhancedNC}, which represents state-of-the-art unstructured pruning methods through the use of tensor decomposition.


\subsection{Experiments on CIFAR-10}
\label{exp:cifar}
\textbf{VGGNet-16.} Table~\ref{table:main} presents our experiments on CIFAR-10 using VGGNet-16. Our methods, sGLP-IB and sTLP-IB, demonstrate strong performance, achieving  accuracy of 94.05\% and 94.10\%, and pruning ratios of 61\% for FLOPs and 85\% for parameters. sTLP-IB achieves the highest Top-1 accuracy among all previous baselines with lower computational overhead and memory requirements, which is 0.14\% higher than the pretrained model and demonstrates the effectiveness of utilizing structured class-wise information. Although sGLP-IB's Top-1 accuracy is slightly lower than AutoBot by 0.05\%, it achieves greater reductions in both FLOPs and parameters. We also present additional results based on varying pruning ratios, showing that our methods remain superior to most baselines even at higher pruning levels. This proves the effectiveness of using graph-structured lasso and tree-guided lasso, which utilize class-wise information, in pruning. Notably, sTLP-IB performs better than sGLP-IB, as the tree-structured approach, which constructs subtrees, captures class information more precisely than graphs.

\noindent \textbf{ResNet-56.} Table~\ref{table:main} also shows our methods, sGLP-IB and sTLP-IB, outperform most baselines using ResNet-56 with Top-1 accuracy rates of 93.93\% and 93.96\%, respectively, while maintaining same pruning ratios of 53\% for FLOPs and 52\% for parameters. sTLP-IB outperforms sGLP-IB, consistent with the results observed with VGGNet-16. Although NORTON slightly surpasses our methods by 0.04\%, its pruning ratios were lower, with only 42\% for FLOPs and 48\% for parameters. Others like DECORE, FTWT, and FSW exhibit higher pruning ratios than SGLP-IB and STLP-IB, but they compromise accuracy. Therefore, our methods achieve a balance between accuracy and pruning ratios, improving upon the pretrained model by 0.68\%.

\begin{table}[!t]
\centering
\resizebox{\linewidth}{!}{
\begin{tabular}{lcccc}
\toprule
\textbf{Method}  &\textbf{Top-1(\%)}    &\textbf{FLOPs}$\downarrow$ &\textbf{Params} $\downarrow$  &  $\Delta$ \textbf{Acc(\%)}   \\
\midrule
\texttt{ResNet-50} & \texttt{76.15}& \texttt{0(4133.74M)} & \texttt{0(25.56M)} & \texttt{0}\\
Hrank	&71.98	&62\%	&62\%	&-4.17\\
DECORE	&72.06	&61\%	&-	&-4.09\\
MFP	    &74.86	&54\%	&-	&-1.29\\
Random	&75.13	&49\%	&54\%	&-1.02\\
SCOP	&75.26	&55\%	&-	&-0.89\\
APIB	&75.37	&62\%	&58\%	&-0.78\\
DPFPS	&75.55	&46\%	&-	&-0.60\\
CC	    &75.59	&53\%	&-	&-0.56\\
LRF-60	&75.71	&56\%	&-	&-0.44\\
NORTON	&75.95	&64\%	&59\%	&-0.20\\
NPPM	&75.96	&56\%	&-	&-0.19\\
CSD	    &76.11	&54\%	&41\%	&-0.04\\
\textbf{sGLP-IB}	&\textbf{76.10}	&\textbf{57\%}	&\textbf{55\%}	&\textbf{-0.05}\\
\textbf{sTLP-IB}	&\textbf{76.12}	&\textbf{57\%}	&\textbf{55\%}	&\textbf{-0.03}\\
\bottomrule
\end{tabular}
}
\vspace{-0.5em}
\caption{Pruning results of ResNet-50 on ImageNet dataset.}
\vspace{-1.5em}
\label{table:imagenet}
\end{table}

\noindent \textbf{ResNet-110.} Both sGLPIB and sTLP-IB achieve state-of-the-art performance on ResNet-110. sTLP-IB behaves better with accuracy of 93.80\%, surpassing the best baseline by 0.12\%, while reducing 62\% of FLOPs and 68\% of parameters as Table~\ref{table:main}, showing an improvement of 0.30\% over the unpruned model. Notably, even though we choose the same pruning ratios, sGLP-IB eliminates more FLOPs and also surpasses all baselines with an accuracy of 93.69\%, albeit 0.11\% lower than sTLP-IB. This further underscores the effectiveness of tree-guided lasso, which retains more critical filters, explaining the FLOPs difference compared to sGLP-IB. Our experiments on ResNet-56 and ResNet-110 validate the efficacy of our pruning methods on ResNet architectures.

\noindent \textbf{GoogleNet.} On GoogleNet, our methods also demonstrate effectiveness, with sGLP-IB reaching 94.75\% and sTLP-IB achieving a test accuracy of 94.78\%. Despite removing 69\% of FLOPs and 63\% of parameters, we significantly reduce overhead and memory size. However, due to the Inception structure of GoogleNet, more fine-grained pruning is necessary. EPruner, which employs Affinity Propagation for fine-grained clustering, achieves better results but with a larger computational overhead compared to ours. Our methods surpass the remaining baselines in terms of accuracy, FLOPs pruning ratio, and parameter pruning ratio. The integration of structured class-wise information with our methods and other approaches holds substantial promise for future advancements.

\subsection{Experiments on ImageNet} 

Compared to the CIFAR dataset, as shown in Table~\ref{table:imagenet}, we conduct experiments on a larger dataset, ImageNet, with numerous baselines using ResNet-50. The results indicate that our sTLP-IB achieves state-of-the-art performance, outperforming all baselines with a Top-1 accuracy of 76.12\%, a 57\% reduction in FLOPs, and a 55\% reduction in parameters, with a mere 0.03\% drop compared to the pretrained model. Our sGLP-IB also outperforms the majority of baselines and was comparable to CSD. Although sGLP-IB achieves a higher pruning ratio than CSD, its accuracy is only 0.01\% lower, which may be attributed to CSD benefiting from its Discrete Wavelet Transform on large datasets, which again shows the effectiveness of our sGLP-IB. 
Combined with the results on CIFAR-10 from Section~\ref{exp:cifar}, our methods are applicable not only to simple datasets but also to complex, real-world datasets, showcasing their significant potential. We can leverage class-wise information and structured lasso for pruning across datasets

\begin{table}[t]
\centering
\resizebox{0.8\linewidth}{!}{
\begin{tabular}{lccccc}
\toprule
\textbf{Method}  &64  & 128 & 256 & 512 & 1024\\
\midrule
sGLP-IB &93.77 &94.05 &94.07 &94.08 & 94.08\\
sTLP-IB &93.76 &94.10 &94.11 &94.10 & 94.11\\
\bottomrule
\end{tabular}
}
\vspace{-0.5em}
\caption{Top-1 Accuracy of our methods using VGGNet-16 on CIFAR-10 dataset when varying the batch sizes.}
\vspace{-0.5em}
\label{table:batchsize}
\end{table}

\begin{table}[t]
\centering
\resizebox{0.8\linewidth}{!}{
\begin{tabular}{lc}
\toprule
  & \textbf{Details of the kernel functions}\\
\midrule
Linear   &$K(x,y)=x^Ty+c$\\
Gaussian &$K(x,y)=exp(\frac{||x-y||^2}{2\sigma^2})$ \\
Sigmoid  &$K(x,y)=tanh(ax^T+c)$ \\
Laplacian&$K(x,y)=exp(\frac{||x-y||}{\sigma})$ \\
\bottomrule
\end{tabular}
}
\vspace{-0.7em}
\caption{Kernel functions of Gram matrix.}
\vspace{-1.5em}
\label{table:kernel}
\end{table}

\begin{table}[t]
\centering
\resizebox{0.8\linewidth}{!}{
\begin{tabular}{lccccc}
\toprule
\textbf{Method}  &Linear  & Gaussian & Sigmoid & Laplacian\\
\midrule
sGLP-IB &94.05  & 94.06& 93.71 & 94.05\\
sTLP-IB &94.07  & 94.09& 93.86 & 94.10\\
\bottomrule
\end{tabular}
}
\vspace{-0.5em}
\caption{Top-1 Accuracy of our methods using VGGNet-16 on CIFAR-10 dataset when varying the kernel functions.}
\vspace{-1.5em}
\label{table:kernel2}
\end{table}




\subsection{Ablation Studies}

\textbf{Influence of batch sizes.} As we keep the same pruning ratio as Table~\ref{table:main}, the accuracy of sGLP-IB and sTLP-IB increases with the batch size but essentially converges once the batch size reaches 128 in Table~\ref{table:batchsize}. This observation forms the basis for setting our experimental batch size to 128, and also ensures a fair comparison with other methods. With a batch size of 64, the randomness due to smaller data volume results in sGLP-IB slightly outperforming sTLP-IB by 0.01\% in Top-1 accuracy. As mentioned in work~\cite{Lin2020HRankFP}, the average rank of feature maps generated by a single filter remains consistently the same and the pruned model's accuracy is not significantly affected by the batch size during the pruning process. Nonetheless, sTLP-IB consistently outperforms sGLP-IB across varying batch sizes, aligning with our previous findings that tree-guided lasso is more effective than graph-structured lasso when the batch size is larger. The experimental results demonstrate that our methods are robust to batch size variations, provided that the random selection of samples within batches is maintained.

\noindent \textbf{Influence of Gram matrix.} Following previous settings, we conduct experiments on the influence of Gram matrix. With the same pruning ratio for VGGNet-16 on CIFAR-10 as Table~\ref{table:main}, sGLP-IB and sTLP-IB achieve Top-1 accuracies of 94.06\% and 94.10\%, respectively. This demonstrates that our methods can be effectively combined with Gram matrix. The integration of Gram matrix with structured lasso accelerates computation without sacrificing class-wise statistical information and accuracy, as evidenced by the comparable results of 94.05\% and 94.10\% obtained with Gram matrix. Notably, the time required for pruning the pretrained model without Gram matrix is approximately 25.7 times greater than when it is used, indicating that the introduction of Gram matrix significantly enhances pruning efficiency. 

\noindent \textbf{Influence of kernel selections.} With the same pruning ratio, we conduct experiments by selecting different kernel functions, as shown in Table~\ref{table:kernel}. Our methods yield similar results across Linear, Gaussian, and Laplacian kernels, maintaining high Top-1 accuracy shown in Table~\ref{table:kernel2}. By utilizing Gram matrix, our methods achieve faster computation and obtain good performance. In all cases, sTLP-IB outperforms sGLP-IB. It is worth noting that both sGLP-IB and sTLP-IB perform slightly worse with the Sigmoid kernel, due to the strong nonlinear properties introduced by the tanh function. But sGLP-IB and sTLP-IB still behave better than most baselines with Sigmoid kernel. Overall, our methods exhibit robustness to different kernel functions.

\begin{figure}[!t]
    \centering
    \subfloat[Linear (T)]{
        \includegraphics[width=0.31\linewidth]{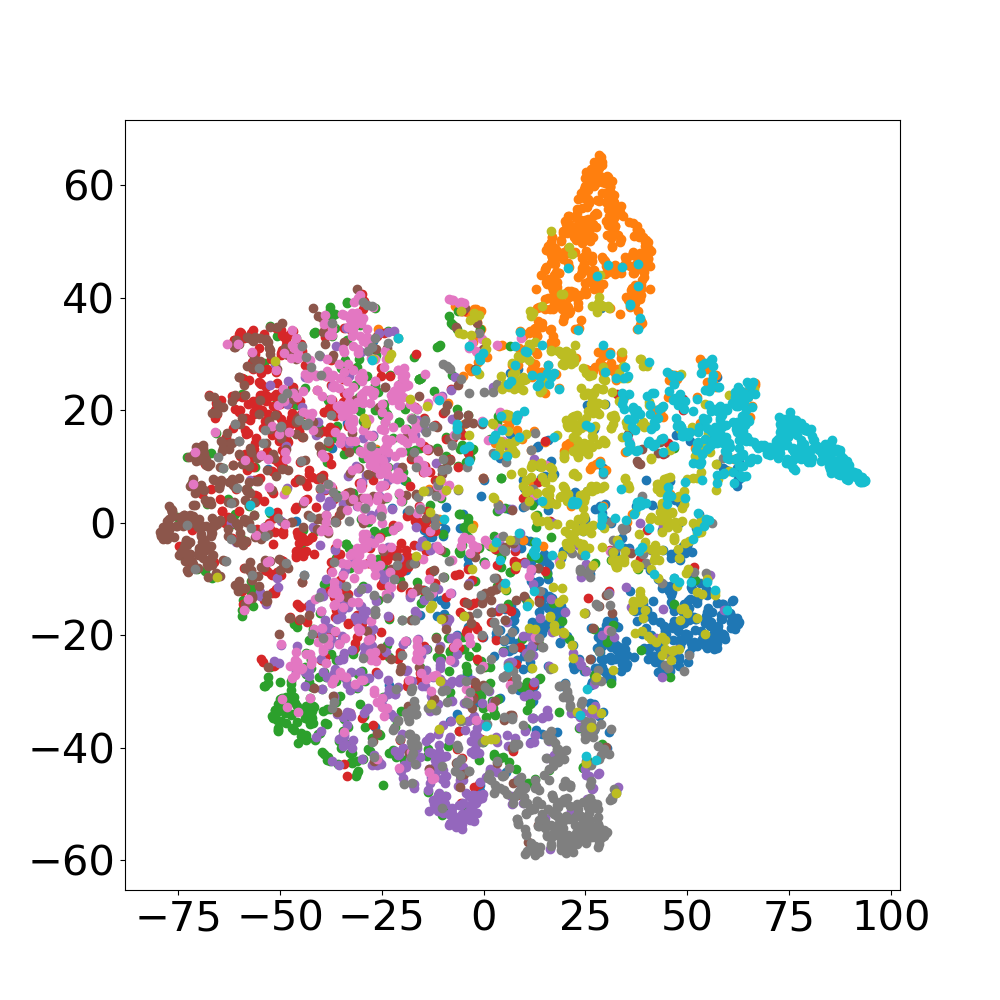}
        \label{fig:lineartree}

    }
    \hfill
    \subfloat[Linear (T)]{
        \includegraphics[width=0.31\linewidth]{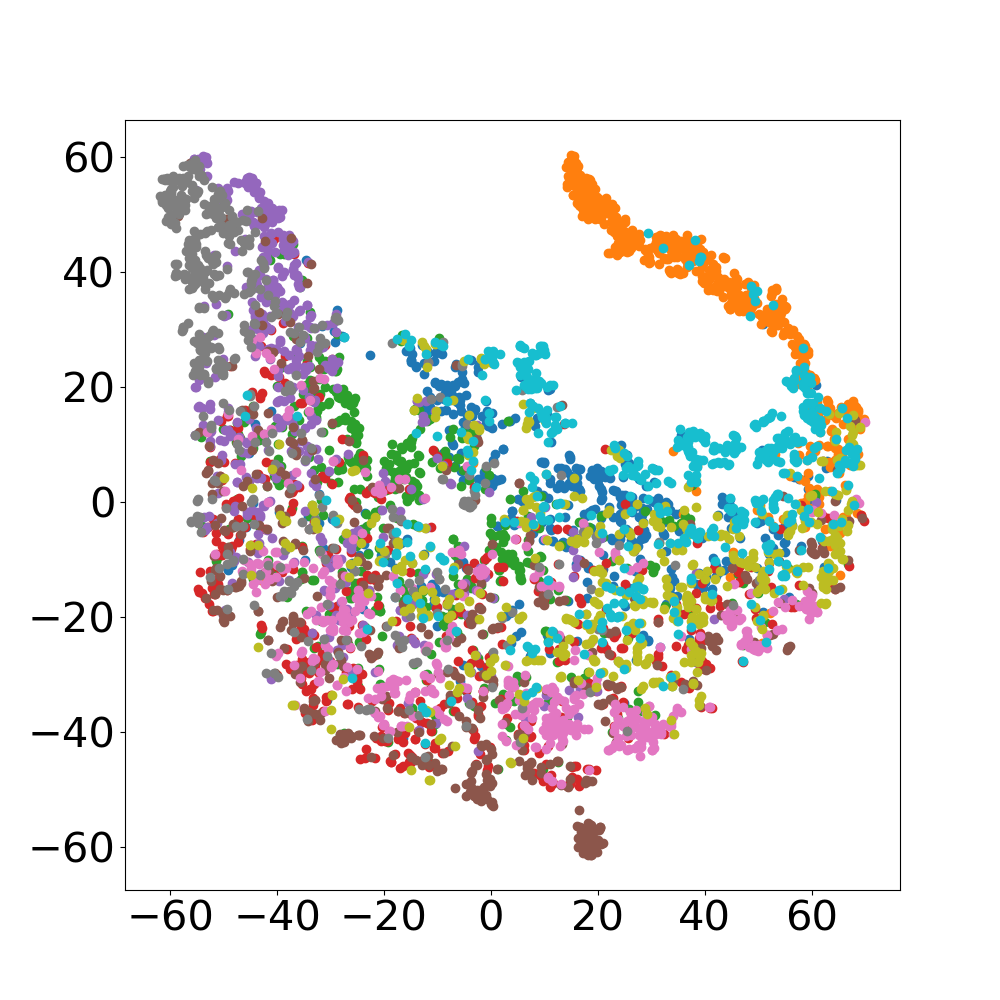}
        \label{fig:lineargraph}
    }
        \hfill
    \subfloat[APIB]{
        \includegraphics[width=0.31\linewidth]{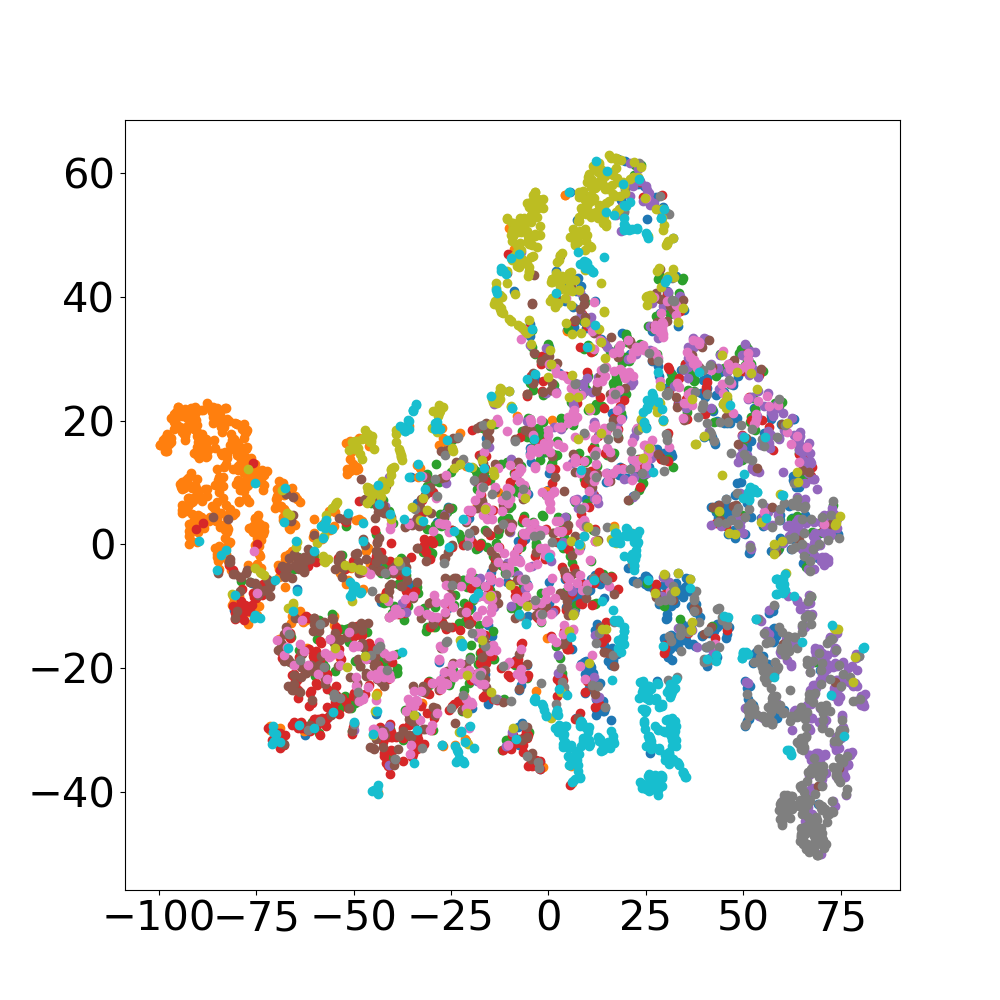}
        \label{fig:linearapib}
    }
    \vspace{-1em}
    \\
    \subfloat[Gaussian (G)]{
        \includegraphics[width=0.31\linewidth]{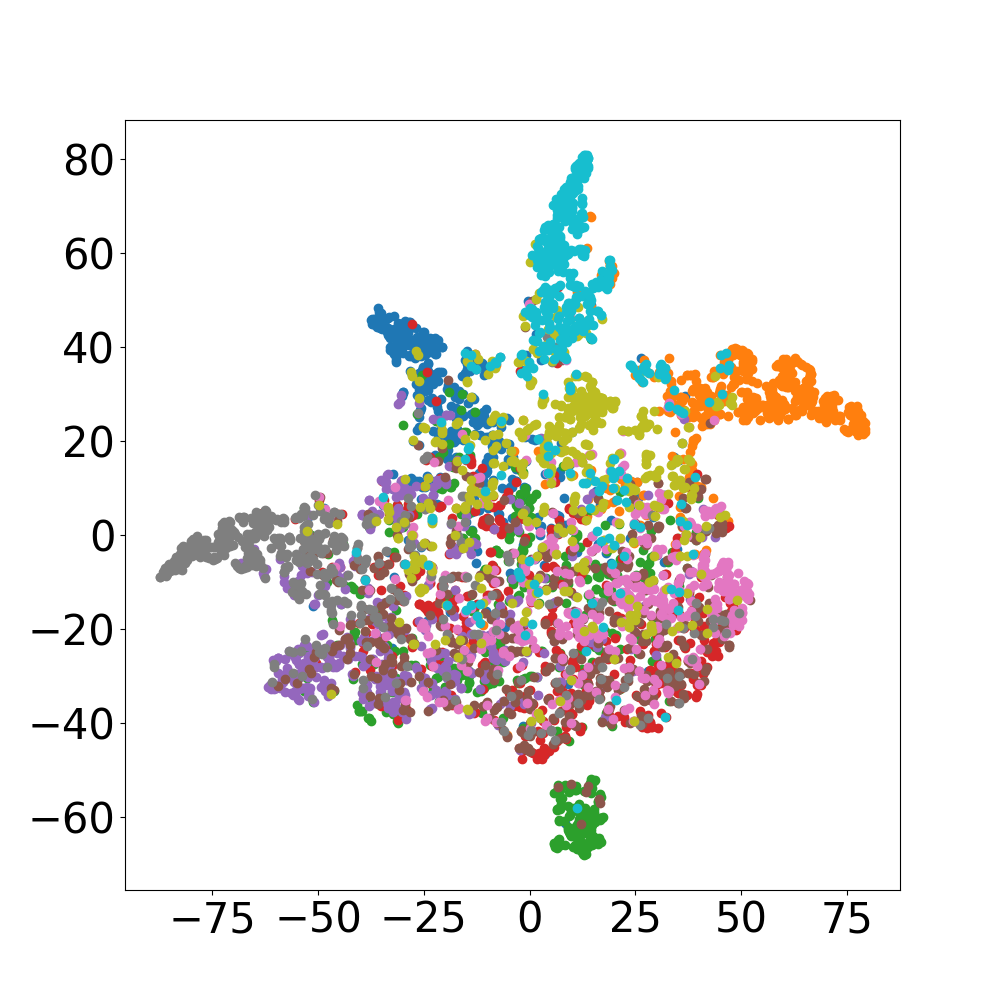}
        \label{fig:gaussiangraph}
    }
    \hfill
    \subfloat[Sigmoid (G)]{
        \includegraphics[width=0.31\linewidth]{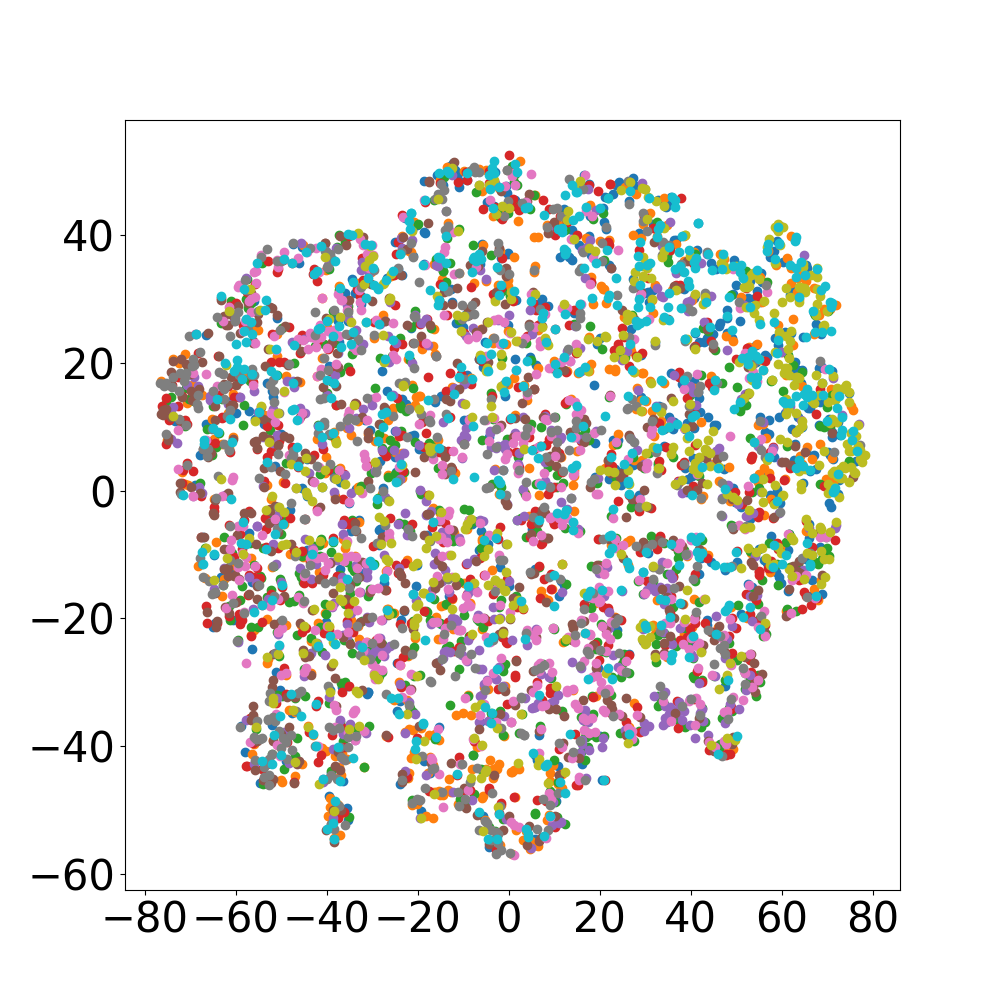}
        \label{fig:sigmoidgraph}
    }
    \hfill
        \subfloat[Laplacian (G)]{
        \includegraphics[width=0.31\linewidth]{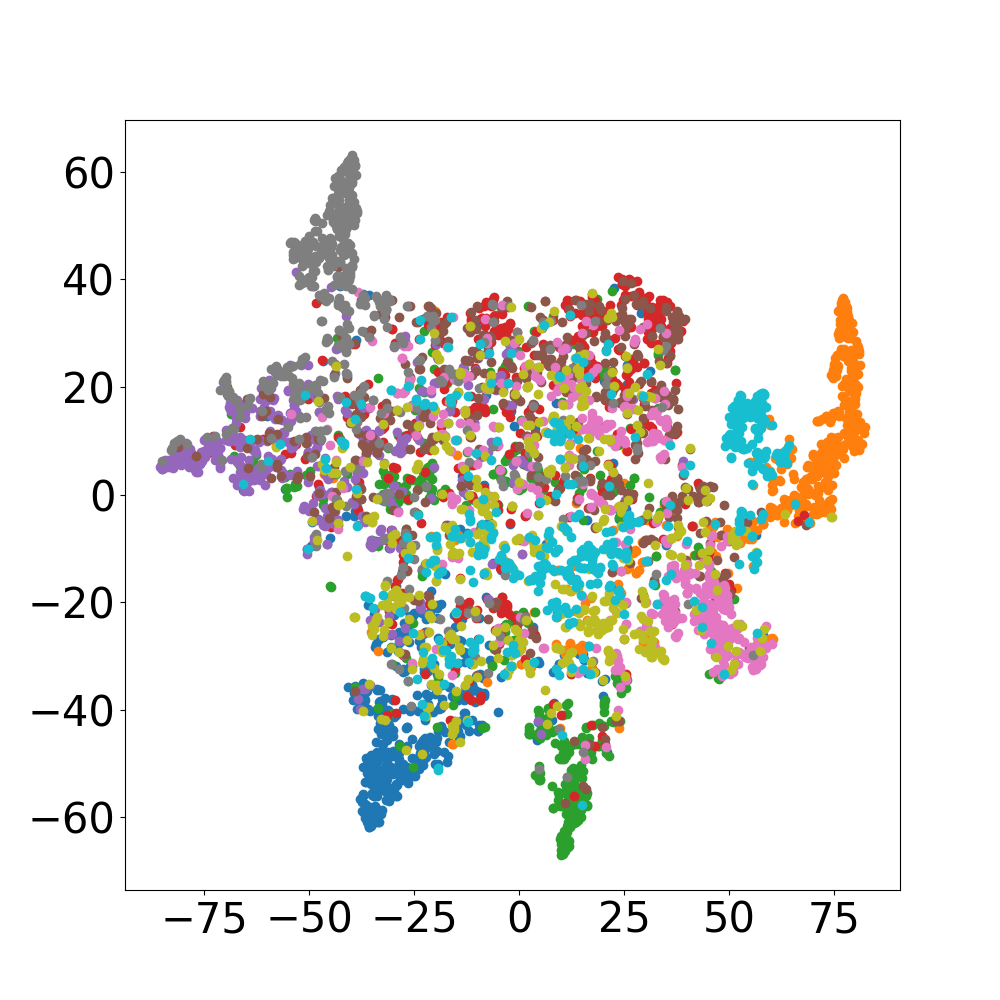}
        \label{fig:laplacegraph}
    }
    \vspace{-1em}
    \\
    \subfloat[Gaussian (T)]{
        \includegraphics[width=0.31\linewidth]{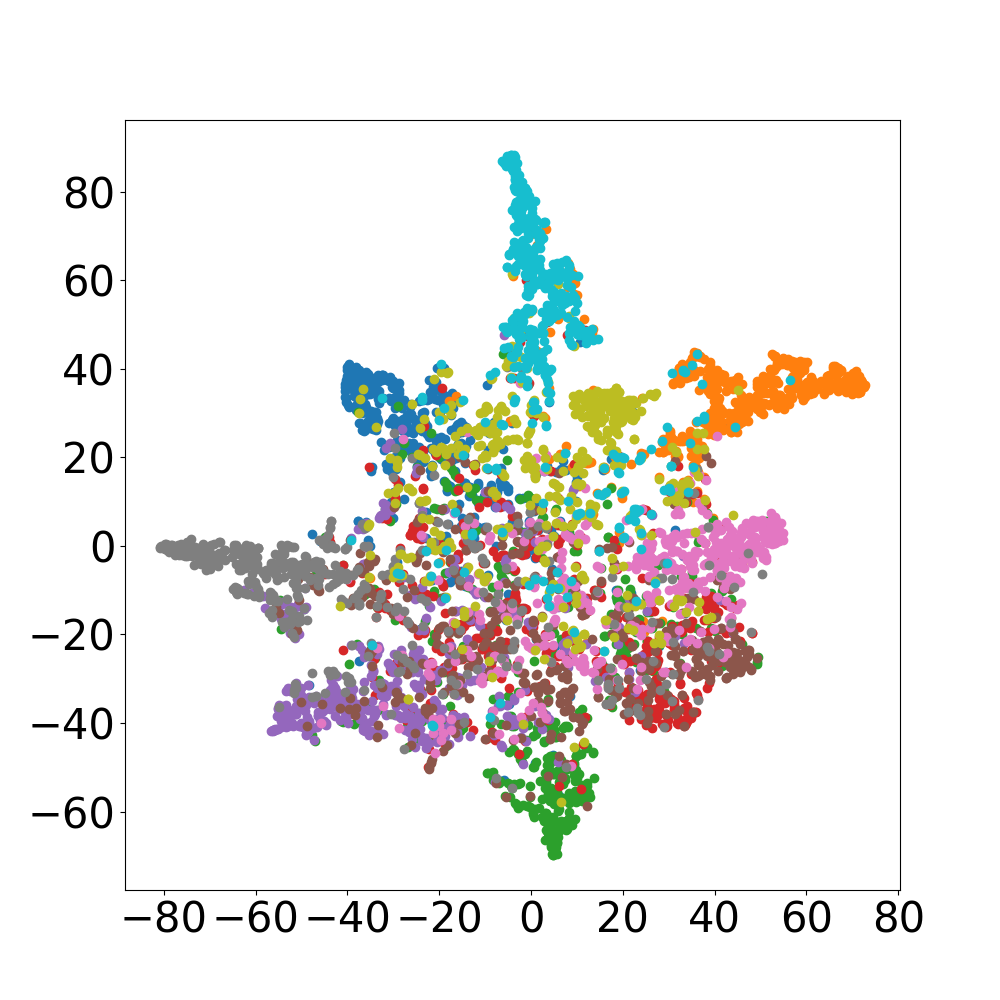}
        \label{fig:gaussiantree}
    }
        \hfill
    \subfloat[Sigmoid (T)]{
        \includegraphics[width=0.31\linewidth]{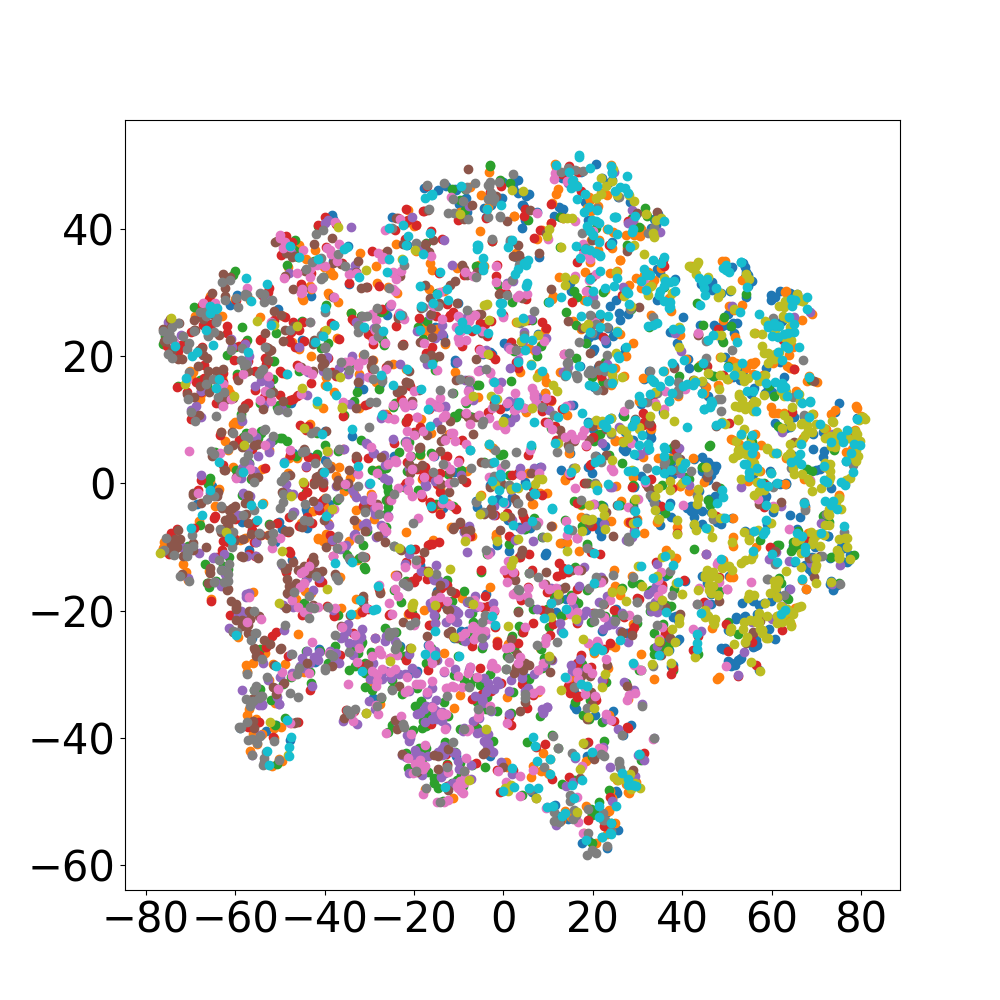}
        \label{fig:sigmoidtree}
    }
    \hfill
    \subfloat [Laplacian (T)]{
        \includegraphics[width=0.31\linewidth]{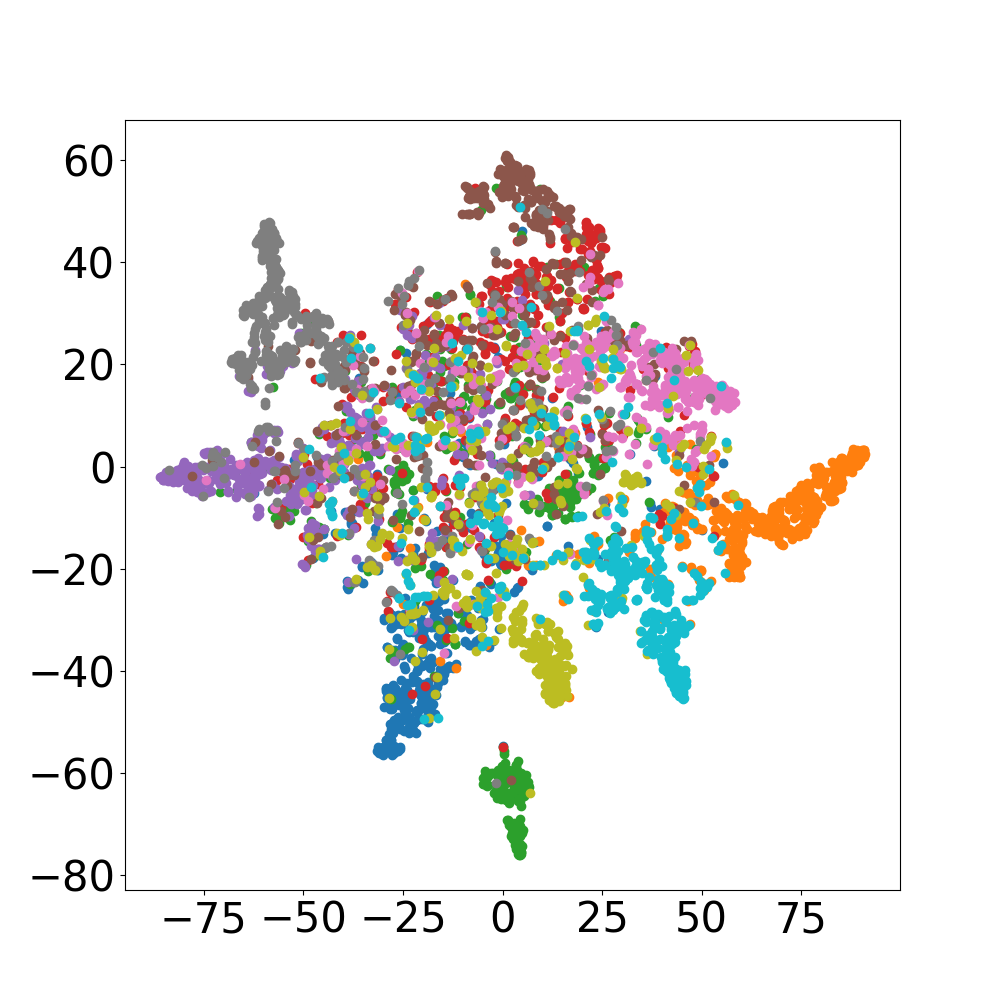}
        \label{fig:laplacetree}
    }
\vspace{-0.5em}
    \caption{t-SNE visualizations of pruned models with different kernel functions before finetuning using VGGNet-16 on CIFAR-10 (T denotes sTLP-IB and G denotes sGLP-IB).}
    \vspace{-1em}
    \label{fig:tsne}
\end{figure}

\begin{table}[!t]
\centering
\resizebox{0.85\linewidth}{!}{
\begin{tabular}{lccccc}
\toprule
\textbf{Method}  &Linear  & Gaussian & Sigmoid & Laplacian\\
\midrule
sGLP-IB &24.85  & 37.90& 10.00 & 42.51\\
sTLP-IB &25.37  & 42.80& 11.13 & 43.68\\
\bottomrule
\end{tabular}
}
\vspace{-0.5em}
\caption{Top-1 Accuracy of our methods when varying the kernel functions without finetuning.}
\vspace{-1.5em}
\label{table:kernel3}
\end{table}

\noindent \textbf{t-SNE of the pruned model before finetuning.} We further analyze the effectiveness of our method using t-SNE visualization on the pruned model before fine-tuning, following the previous settings. As illustrated in Fig.~\ref{fig:tsne}, we also compare our approach with the most similar state-of-the-art method, APIB. Under the linear function, both sGLP-IB and sTLP-IB from our method perform better than APIB (Row 1), with more distinct clusters between classes, indicating that our approach better leverages structured class-wise information. We also present t-SNE visualizations for the Gaussian, Sigmoid, and Laplacian kernels (Row 2 and Row 3). We find that both the Gaussian and Laplacian kernels aid in learning class information, while the Sigmoid kernel function results in almost random distributions, consistent with the results in Table~\ref{table:kernel2}.
We also evaluate the pruned model's Top-1 accuracy before finetuning, as shown in Table~\ref{table:kernel3}. We observe that the Gaussian and Laplacian kernels facilitate learning more information in high-dimensional spaces, while the linear kernel demonstrates the inherent capability of our structured lasso. The results with the Sigmoid kernel mirror the t-SNE visualizations. Notably, both in Table~\ref{table:kernel3} and Fig.~\ref{fig:tsne}, sTLP-IB consistently outperformed sGLP-IB, which aligns with the results in Table~\ref{table:main}.

\begin{table}[!t]
\centering
\resizebox{\linewidth}{!}{
\begin{tabular}{lcccccc}
\toprule
\textbf{Ratio}  &10  & 30 & 50 & 70 & 90\\
\midrule
sGLP-IB & 94.10(8\%) & 94.13(32\%)& 94.10(54\%) & 94.08(63\%) & 93.97(64\%)\\
sTLP-IB & 94.10(8\%) & 94.13(32\%)& 94.11(54\%) & 94.10(63\%) & 94.01(64\%)\\
\bottomrule
\end{tabular}
}
\vspace{-0.5em}
\caption{Results of our methods when varying the parameter pruning ratios using VGGNet-16 on CIFAR-10 (The pruning ratio of pruned FLOPs is indicated in parentheses.).}
\vspace{-1.5em}
\label{table:ratio}
\end{table}



\noindent \textbf{Influence of pruning ratio for Parameters.} By adjusting the pruning ratio in Table~\ref{table:ratio}, we find that when the ratio is below 50\%, the performance of sGLP-IB and sTLP-IB is similar due to the low pruning rate, resulting in nearly identical retained information and thus similar accuracy. When the ratio is below 70\%, the variations in accuracy are minimal, indicating that our methods effectively capture and utilize statistical information for pruning. When the ratio falls below 10\%, the accuracy has a slight drop, due to the excessive number of redundant filters.


\noindent \textbf{Influence of layer pruning.} We conduct experiments to assess effects of retaining the initial few layers without pruning. To maintain the same pruning ratio, we compensate by pruning more in later layers. In this scenario, accuracies for sGLP-IB and sTLP-IB are 93.76\% and 93.88\% on CIFAR-10 with VGGNet-16, respectively. If the first two layers were not pruned, the accuracies are 93.79\% and 93.98\%, respectively. These results are lower than those in Table~\ref{table:main}, which can be attributed to the fact that early layers focus on low-level features of inputs, while later layers contain more structured class-wise information.





\subsection{Supplementary Materials} 
Artifact evaluation details, time complexity and convergence discussions, adaption for pruning ResNet and GoogleNet, dataset details, experiments with CIFAR-100 dataset, experiments with pruning ratios for FLOPs, real implementation on resource-constrained devices and our resource-constrained device prototype can be found in Appendix.

\section{Discussions}
\label{sec:discussions}
Our paper demonstrates the effectiveness of utilizing structured class-wise information. Although this work focuses on classification, our method can also be applied to segmentation and object detection in the future, as filters likewise exhibit structured class information. Therefore, exploring better methods to capture more precise class-wise information is one of our future directions. Our method is orthogonal to quantization approaches~\cite{li2023repq,li2023vit}, and can be combined with them in the future to further reduce computational overhead.
Moreover, as Transformer~\cite{vaswani2017attention} architectures are increasingly adopted in multimedia applications and their large model sizes pose challenges for deployment on resource-constrained devices, we plan to explore applying our effective method to Transformer-based models as our future research focus to explore the effectiveness of our methods.

\section{Conclusion}
\label{sec:conclusion}
We propose two novel automatic pruning methods that both achieve state-of-the-art, sGLP-IB and sTLP-IB, which leverage information bottleneck theory and structured lasso to perform pruning using penalties that construct subgroup and subtree structures based on class-wise statistical information. Additionally, we present an adaptive approach that utilizes Gram matrix to accelerate the process and employs proximal gradient descent to further ensure convergence and polynomial time complexity. Through extensive experiments and ablation studies, we demonstrate the potential of our methods—both theoretically and experimentally—to effectively leverage structured class-wise information for pruning tasks and facilitate deployment on resource-constrained devices, compared with numerous baselines.


\begin{thebibliography}{60}
\providecommand{\natexlab}[1]{#1}

\bibitem[{Alwani, Madhavan, and Wang(2022)}]{Alwani2021DECOREDC}
Alwani, M.; Madhavan, V.; and Wang, Y. 2022.
\newblock DECORE: Deep Compression with Reinforcement Learning.
\newblock \emph{IEEE/CVF Conference on Computer Vision and Pattern Recognition}.

\bibitem[{Cai et~al.(2022)Cai, An, Yang, Yan, and Xu}]{Cai2022PriorGM}
Cai, L.; An, Z.; Yang, C.; Yan, Y.; and Xu, Y. 2022.
\newblock Prior Gradient Mask Guided Pruning-Aware Fine-Tuning.
\newblock In \emph{AAAI}.

\bibitem[{Castells and Yeom(2021)}]{Castells2021AutomaticNN}
Castells, T.; and Yeom, S.-K. 2021.
\newblock Automatic Neural Network Pruning that Efficiently Preserves the Model Accuracy.
\newblock \emph{ArXiv}, abs/2111.09635.

\bibitem[{Chen et~al.(2010)Chen, Kim, Lin, Carbonell, and Xing}]{Chen2010GraphStructuredMR}
Chen, X.; Kim, S.; Lin, Q.; Carbonell, J.~G.; and Xing, E.~P. 2010.
\newblock Graph-Structured Multi-task Regression and an Efficient Optimization Method for General Fused Lasso.
\newblock \emph{ArXiv}, abs/1005.3579.

\bibitem[{Chen et~al.(2012)Chen, Lin, Kim, Carbonell, and Xing}]{chen2012smoothing}
Chen, X.; Lin, Q.; Kim, S.; Carbonell, J.~G.; and Xing, E.~P. 2012.
\newblock Smoothing proximal gradient method for general structured sparse regression.
\newblock \emph{The Annals of Applied Statistics}, 719--752.

\bibitem[{Dai, Zhu, and Wipf(2018)}]{Dai2018CompressingNN}
Dai, B.; Zhu, C.; and Wipf, D.~P. 2018.
\newblock Compressing Neural Networks using the Variational Information Bottleneck.
\newblock In \emph{International Conference on Machine Learning}.

\bibitem[{Deng et~al.(2009)Deng, Dong, Socher, Li, Li, and Fei-Fei}]{deng2009imagenet}
Deng, J.; Dong, W.; Socher, R.; Li, L.-J.; Li, K.; and Fei-Fei, L. 2009.
\newblock Imagenet: A large-scale hierarchical image database.
\newblock In \emph{2009 IEEE conference on computer vision and pattern recognition}, 248--255. Ieee.

\bibitem[{Ding et~al.(2019)Ding, Ding, Zhou, Guo, Liu, and Han}]{Ding2019GlobalSM}
Ding, X.; Ding, G.; Zhou, X.; Guo, Y.; Liu, J.; and Han, J. 2019.
\newblock Global Sparse Momentum SGD for Pruning Very Deep Neural Networks.
\newblock \emph{Advances in Neural Information Processing Systems}.

\bibitem[{Duan et~al.(2022)Duan, Hu, Zhou, He, Liu, and Duan}]{Duan2022NetworkPV}
Duan, Y.; Hu, X.; Zhou, Y.; He, P.; Liu, Q.; and Duan, S. 2022.
\newblock Network Pruning via Feature Shift Minimization.
\newblock In \emph{Asian Conference on Computer Vision}.

\bibitem[{Elkerdawy et~al.(2022)Elkerdawy, Elhoushi, Zhang, and Ray}]{Elkerdawy2021FireTW}
Elkerdawy, S.; Elhoushi, M.; Zhang, H.; and Ray, N. 2022.
\newblock Fire Together Wire Together: A Dynamic Pruning Approach with Self-Supervised Mask Prediction.
\newblock \emph{CVPR}.

\bibitem[{Gao et~al.(2021)Gao, Huang, Cai, and Huang}]{Gao2021NetworkPV}
Gao, S.; Huang, F.; Cai, W.~T.; and Huang, H. 2021.
\newblock Network Pruning via Performance Maximization.
\newblock \emph{IEEE/CVF Conference on Computer Vision and Pattern Recognition}.

\bibitem[{Guo et~al.(2023)Guo, Zhang, Zheng, Wang, Li, Chao, Wu, Zhang, and Ji}]{guo2023automatic}
Guo, S.; Zhang, L.; Zheng, X.; Wang, Y.; Li, Y.; Chao, F.; Wu, C.; Zhang, S.; and Ji, R. 2023.
\newblock Automatic network pruning via hilbert-schmidt independence criterion lasso under information bottleneck principle.
\newblock In \emph{Proceedings of the IEEE/CVF international conference on computer vision}.

\bibitem[{Guo et~al.(2021)Guo, Yuan, Tan, Wang, Yang, and Liu}]{Guo2021GDPSN}
Guo, Y.; Yuan, H.; Tan, J.; Wang, Z.; Yang, S.; and Liu, J. 2021.
\newblock GDP: Stabilized Neural Network Pruning via Gates with Differentiable Polarization.
\newblock \emph{IEEE/CVF International Conference on Computer Vision}.

\bibitem[{He et~al.(2016)He, Zhang, Ren, and Sun}]{He2015DeepRL}
He, K.; Zhang, X.; Ren, S.; and Sun, J. 2016.
\newblock Deep Residual Learning for Image Recognition.
\newblock \emph{IEEE Conference on Computer Vision and Pattern Recognition}.

\bibitem[{He et~al.(2018)He, Liu, Wang, Hu, and Yang}]{He2018FilterPV}
He, Y.; Liu, P.; Wang, Z.; Hu, Z.; and Yang, Y. 2018.
\newblock Filter Pruning via Geometric Median for Deep Convolutional Neural Networks Acceleration.
\newblock \emph{2019 IEEE/CVF Conference on Computer Vision and Pattern Recognition (CVPR)}, 4335--4344.

\bibitem[{He et~al.(2019)He, Liu, Zhu, and Yang}]{He2019FilterPB}
He, Y.; Liu, P.; Zhu, L.; and Yang, Y. 2019.
\newblock Filter Pruning by Switching to Neighboring CNNs With Good Attributes.
\newblock \emph{IEEE Transactions on Neural Networks and Learning Systems}, 34: 8044--8056.

\bibitem[{He, Zhang, and Sun(2017)}]{He2017ChannelPF}
He, Y.; Zhang, X.; and Sun, J. 2017.
\newblock Channel Pruning for Accelerating Very Deep Neural Networks.
\newblock \emph{ICCV}.

\bibitem[{He et~al.(2022)He, Qian, Wang, Wang, Guan, Gu, Ling, Zeng, Wang, and Zhou}]{He2022FilterPV}
He, Z.; Qian, Y.; Wang, Y.; Wang, B.; Guan, X.; Gu, Z.; Ling, X.; Zeng, S.; Wang, H.; and Zhou, W. 2022.
\newblock Filter Pruning via Feature Discrimination in Deep Neural Networks.
\newblock In \emph{European Conference on Computer Vision}.

\bibitem[{Hemmat, San~Miguel, and Davoodi(2020)}]{hemmat2020cap}
Hemmat, M.; San~Miguel, J.; and Davoodi, A. 2020.
\newblock CAP’NN: Class-aware personalized neural network inference.
\newblock In \emph{2020 57th ACM/IEEE Design Automation Conference (DAC)}, 1--6. IEEE.

\bibitem[{Hu et~al.(2016)Hu, Peng, Tai, and Tang}]{Hu2016NetworkTA}
Hu, H.; Peng, R.; Tai, Y.-W.; and Tang, C.-K. 2016.
\newblock Network Trimming: A Data-Driven Neuron Pruning Approach towards Efficient Deep Architectures.
\newblock \emph{ArXiv}, abs/1607.03250.

\bibitem[{Hu et~al.(2023)Hu, Che, Liu, Li, Tang, Zhang, and Wang}]{hu2023catro}
Hu, W.; Che, Z.; Liu, N.; Li, M.; Tang, J.; Zhang, C.; and Wang, J. 2023.
\newblock CATRO: Channel pruning via class-aware trace ratio optimization.
\newblock \emph{IEEE Transactions on Neural Networks and Learning Systems}, 35(8): 11595--11607.

\bibitem[{Hussien et~al.(2024)Hussien, Afifi, Nguyen, and Cheriet}]{Hussien2024SmallCS}
Hussien, M.; Afifi, M.; Nguyen, K.~K.; and Cheriet, M. 2024.
\newblock Small Contributions, Small Networks: Efficient Neural Network Pruning Based on Relative Importance.
\newblock \emph{ArXiv}, abs/2410.16151.

\bibitem[{Joo et~al.(2021)Joo, Yi, Baek, and Kim}]{Joo2021LinearlyRF}
Joo, D.; Yi, E.; Baek, S.; and Kim, J. 2021.
\newblock Linearly Replaceable Filters for Deep Network Channel Pruning.
\newblock In \emph{AAAI}.

\bibitem[{Kim and Xing(2009)}]{Kim2009TreeGuidedGL}
Kim, S.; and Xing, E.~P. 2009.
\newblock Tree-Guided Group Lasso for Multi-Task Regression with Structured Sparsity.
\newblock In \emph{International Conference on Machine Learning}.

\bibitem[{Krizhevsky(2009)}]{Krizhevsky2009LearningML}
Krizhevsky, A. 2009.
\newblock Learning Multiple Layers of Features from Tiny Images.

\bibitem[{Krizhevsky, Sutskever, and Hinton(2012)}]{Krizhevsky2012ImageNetCW}
Krizhevsky, A.; Sutskever, I.; and Hinton, G.~E. 2012.
\newblock ImageNet classification with deep convolutional neural networks.
\newblock \emph{Communications of the ACM}, 60: 84 -- 90.

\bibitem[{Lanckriet et~al.(2002)Lanckriet, Cristianini, Bartlett, Ghaoui, and Jordan}]{Lanckriet2002LearningTK}
Lanckriet, G. R.~G.; Cristianini, N.; Bartlett, P.~L.; Ghaoui, L.~E.; and Jordan, M.~I. 2002.
\newblock Learning the Kernel Matrix with Semidefinite Programming.
\newblock In \emph{Journal of machine learning research}.

\bibitem[{Lee et~al.(2020)Lee, Lee, Lee, and Song}]{Lee2020ChannelPV}
Lee, M.~K.; Lee, S.; Lee, S.~H.; and Song, B.~C. 2020.
\newblock Channel Pruning Via Gradient Of Mutual Information For Light-Weight Convolutional Neural Networks.
\newblock \emph{2020 IEEE International Conference on Image Processing (ICIP)}, 1751--1755.

\bibitem[{Li et~al.(2016)Li, Kadav, Durdanovic, Samet, and Graf}]{Li2016PruningFF}
Li, H.; Kadav, A.; Durdanovic, I.; Samet, H.; and Graf, H.~P. 2016.
\newblock Pruning Filters for Efficient ConvNets.
\newblock \emph{ArXiv}, abs/1608.08710.

\bibitem[{Li et~al.(2022)Li, Adamczewski, Li, Gu, Timofte, and Gool}]{Li2022RevisitingRC}
Li, Y.; Adamczewski, K.; Li, W.; Gu, S.; Timofte, R.; and Gool, L.~V. 2022.
\newblock Revisiting Random Channel Pruning for Neural Network Compression.
\newblock \emph{CVPR}.

\bibitem[{Li et~al.(2021)Li, Lin, Liu, Ye, Wang, Chao, Yang, Ma, Tian, and Ji}]{Li2021TowardsCC}
Li, Y.; Lin, S.; Liu, J.; Ye, Q.; Wang, M.; Chao, F.; Yang, F.; Ma, J.; Tian, Q.; and Ji, R. 2021.
\newblock Towards Compact CNNs via Collaborative Compression.
\newblock \emph{CVPR}.

\bibitem[{Li and Gu(2023)}]{li2023vit}
Li, Z.; and Gu, Q. 2023.
\newblock I-vit: Integer-only quantization for efficient vision transformer inference.
\newblock In \emph{Proceedings of the IEEE/CVF International Conference on Computer Vision}, 17065--17075.

\bibitem[{Li et~al.(2023)Li, Xiao, Yang, and Gu}]{li2023repq}
Li, Z.; Xiao, J.; Yang, L.; and Gu, Q. 2023.
\newblock Repq-vit: Scale reparameterization for post-training quantization of vision transformers.
\newblock In \emph{Proceedings of the IEEE/CVF International Conference on Computer Vision}, 17227--17236.

\bibitem[{Lin et~al.(2021)Lin, Ji, Li, Wang, Wu, Huang, and Ye}]{Lin2021NetworkPU}
Lin, M.; Ji, R.; Li, S.; Wang, Y.; Wu, Y.; Huang, F.; and Ye, Q. 2021.
\newblock Network Pruning Using Adaptive Exemplar Filters.
\newblock \emph{IEEE Transactions on Neural Networks and Learning Systems}, 33: 7357--7366.

\bibitem[{Lin et~al.(2020)Lin, Ji, Wang, Zhang, Zhang, Tian, and Shao}]{Lin2020HRankFP}
Lin, M.; Ji, R.; Wang, Y.; Zhang, Y.; Zhang, B.; Tian, Y.; and Shao, L. 2020.
\newblock HRank: Filter Pruning Using High-Rank Feature Map.
\newblock \emph{CVPR}.

\bibitem[{Liu et~al.(2024{\natexlab{a}})Liu, Liu, Diao, Ye, Liu, and Wei}]{Liu2024SparseVS}
Liu, X.; Liu, H.; Diao, J.; Ye, W.; Liu, X.; and Wei, D. 2024{\natexlab{a}}.
\newblock Sparse Variable Selection on High Dimensional Heterogeneous Data With Tree Structured Responses.
\newblock \emph{IEEE Access}.

\bibitem[{Liu et~al.(2024{\natexlab{b}})Liu, Liu, Diao, Zheng, Li, Xie, Lai, Geng, Song, and Jiang}]{Liu2024NovelTG}
Liu, X.; Liu, X.; Diao, J.; Zheng, M.; Li, J.; Xie, Y.; Lai, K.; Geng, X.; Song, Y.; and Jiang, L. 2024{\natexlab{b}}.
\newblock Novel Truncated-rank Graph-structured and Tree-guided Sparse Linear Mixed Models for Variable Selection on Genome-wide Association Studies.
\newblock \emph{BIBM}.

\bibitem[{Liu et~al.(2024{\natexlab{c}})Liu, Song, Li, Sun, Lan, Liu, Jiang, and Li}]{Liu2024EDViTSV}
Liu, X.; Song, Y.; Li, X.; Sun, Y.; Lan, H.; Liu, Z.; Jiang, L.; and Li, J. 2024{\natexlab{c}}.
\newblock ED-ViT: Splitting Vision Transformer for Distributed Inference on Edge Devices.
\newblock \emph{ArXiv}, abs/2410.11650.

\bibitem[{Ma, Lewis, and Kleijn(2020)}]{Ma2019TheHB}
Ma, K. W.-D.; Lewis, J.~P.; and Kleijn, W. 2020.
\newblock The HSIC Bottleneck: Deep Learning without Back-Propagation.
\newblock \emph{AAAI}.

\bibitem[{Ma et~al.(2023)Ma, Zhao, Liu, He, and Zhou}]{ma2023ocap}
Ma, Y.-D.; Zhao, Z.-C.; Liu, D.; He, Z.; and Zhou, W. 2023.
\newblock OCAP: On-device Class-Aware Pruning for personalized edge DNN models.
\newblock \emph{Journal of Systems Architecture}, 142: 102956.

\bibitem[{Pham, Zniyed, and Nguyen(2024{\natexlab{a}})}]{Pham2024EfficientTD}
Pham, V.~T.; Zniyed, Y.; and Nguyen, T.~P. 2024{\natexlab{a}}.
\newblock Efficient tensor decomposition-based filter pruning.
\newblock \emph{Neural networks : the official journal of the International Neural Network Society}, 178: 106393.

\bibitem[{Pham, Zniyed, and Nguyen(2024{\natexlab{b}})}]{Pham2024EnhancedNC}
Pham, V.~T.; Zniyed, Y.; and Nguyen, T.~P. 2024{\natexlab{b}}.
\newblock Enhanced Network Compression Through Tensor Decompositions and Pruning.
\newblock \emph{IEEE transactions on neural networks and learning systems}.

\bibitem[{Rajaraman et~al.(2023)Rajaraman, Devvrit, Mokhtari, and Ramchandran}]{Rajaraman2023GreedyPW}
Rajaraman, N.; Devvrit; Mokhtari, A.; and Ramchandran, K. 2023.
\newblock Greedy Pruning with Group Lasso Provably Generalizes for Matrix Sensing and Neural Networks with Quadratic Activations.
\newblock \emph{NeurIPS}.

\bibitem[{Ruan et~al.(2021)Ruan, Liu, Li, Yuan, and Hu}]{Ruan2021DPFPSDA}
Ruan, X.; Liu, Y.; Li, B.; Yuan, C.; and Hu, W. 2021.
\newblock DPFPS: Dynamic and Progressive Filter Pruning for Compressing Convolutional Neural Networks from Scratch.
\newblock In \emph{AAAI}.

\bibitem[{Sakamoto and Sato(2024)}]{Sakamoto2024EndtoEndTI}
Sakamoto, K.; and Sato, I. 2024.
\newblock End-to-End Training Induces Information Bottleneck through Layer-Role Differentiation: A Comparative Analysis with Layer-wise Training.
\newblock \emph{ArXiv}, abs/2402.09050.

\bibitem[{Scardapane et~al.(2016)Scardapane, Comminiello, Hussain, and Uncini}]{Scardapane2016GroupSR}
Scardapane, S.; Comminiello, D.; Hussain, A.; and Uncini, A. 2016.
\newblock Group sparse regularization for deep neural networks.
\newblock \emph{Neurocomputing}, 241: 81--89.

\bibitem[{Shao and Shin(2022)}]{Shao2022StructuredPF}
Shao, T.; and Shin, D. 2022.
\newblock Structured Pruning for Deep Convolutional Neural Networks via Adaptive Sparsity Regularization.
\newblock \emph{IEEE 46th Annual Computers, Software, and Applications Conference}.

\bibitem[{Simonyan and Zisserman(2014)}]{Simonyan2014VeryDC}
Simonyan, K.; and Zisserman, A. 2014.
\newblock Very Deep Convolutional Networks for Large-Scale Image Recognition.
\newblock \emph{CoRR}, abs/1409.1556.

\bibitem[{Szegedy et~al.(2015)Szegedy, Liu, Jia, Sermanet, Reed, Anguelov, Erhan, Vanhoucke, and Rabinovich}]{Szegedy2014GoingDW}
Szegedy, C.; Liu, W.; Jia, Y.; Sermanet, P.; Reed, S.~E.; Anguelov, D.; Erhan, D.; Vanhoucke, V.; and Rabinovich, A. 2015.
\newblock Going deeper with convolutions.
\newblock \emph{IEEE Conference on Computer Vision and Pattern Recognition}.

\bibitem[{Tang et~al.(2020)Tang, Wang, Xu, Tao, Xu, Xu, and Xu}]{Tang2020SCOPSC}
Tang, Y.; Wang, Y.; Xu, Y.; Tao, D.; Xu, C.; Xu, C.; and Xu, C. 2020.
\newblock SCOP: Scientific Control for Reliable Neural Network Pruning.
\newblock \emph{ArXiv}, abs/2010.10732.

\bibitem[{Tishby, Pereira, and Bialek(2000)}]{Tishby2000TheIB}
Tishby, N.; Pereira, F.~C.; and Bialek, W. 2000.
\newblock The information bottleneck method.
\newblock \emph{ArXiv}, physics/0004057.

\bibitem[{Tishby and Zaslavsky(2015)}]{Tishby2015DeepLA}
Tishby, N.; and Zaslavsky, N. 2015.
\newblock Deep learning and the information bottleneck principle.
\newblock \emph{IEEE Information Theory Workshop}.

\bibitem[{Vaswani et~al.(2017)Vaswani, Shazeer, Parmar, Uszkoreit, Jones, Gomez, Kaiser, and Polosukhin}]{vaswani2017attention}
Vaswani, A.; Shazeer, N.; Parmar, N.; Uszkoreit, J.; Jones, L.; Gomez, A.~N.; Kaiser, {\L}.; and Polosukhin, I. 2017.
\newblock Attention is all you need.
\newblock \emph{Advances in neural information processing systems}, 30.

\bibitem[{Wang, Li, and Wang(2021)}]{Wang2021ConvolutionalNN}
Wang, Z.; Li, C.; and Wang, X. 2021.
\newblock Convolutional Neural Network Pruning with Structural Redundancy Reduction.
\newblock \emph{IEEE/CVF Conference on Computer Vision and Pattern Recognition}.

\bibitem[{Xie et~al.(2024)Xie, Yuan, Ma, and Li}]{Xie2024AdaptivePO}
Xie, W.; Yuan, M.; Ma, J.; and Li, Y. 2024.
\newblock Adaptive Pruning of Channel Spatial Dependability in Convolutional Neural Networks.
\newblock In \emph{ACM Multimedia}.

\bibitem[{Yan et~al.(2021)Yan, Li, Guo, Yang, and Xu}]{Yan2020ChannelPV}
Yan, Y.; Li, C.; Guo, R.; Yang, K.; and Xu, Y. 2021.
\newblock Channel Pruning via Multi-Criteria based on Weight Dependency.
\newblock \emph{2021 International Joint Conference on Neural Networks (IJCNN)}.

\bibitem[{Yu et~al.(2024)Yu, Du, Jiang, Tong, and Deng}]{ijcai2024p596}
Yu, D.; Du, X.; Jiang, L.; Tong, W.; and Deng, S. 2024.
\newblock EC-SNN: Splitting Deep Spiking Neural Networks for Edge Devices.
\newblock In \emph{Proceedings of the Thirty-Third International Joint Conference on Artificial Intelligence}.

\bibitem[{Yuan(2024)}]{Yuan2024SmoothingPG}
Yuan, G. 2024.
\newblock Smoothing Proximal Gradient Methods for Nonsmooth Sparsity Constrained Optimization: Optimality Conditions and Global Convergence.
\newblock In \emph{International Conference on Machine Learning}.

\bibitem[{Zhang, Gao, and Huang(2021)}]{Zhang2021ExplorationAE}
Zhang, Y.; Gao, S.; and Huang, H. 2021.
\newblock Exploration and Estimation for Model Compression.
\newblock \emph{ICCV}.

\bibitem[{Zhang et~al.(2022)Zhang, Lin, Lin, Luo, Li, Chao, Wu, and Ji}]{Zhang2022LearningBC}
Zhang, Y.; Lin, M.; Lin, Z.; Luo, Y.; Li, K.; Chao, F.; Wu, Y.; and Ji, R. 2022.
\newblock Learning Best Combination for Efficient N: M Sparsity.
\newblock \emph{ArXiv}, abs/2206.06662.

\end{thebibliography}

\newpage
\appendix

\section{Artifact Evulation}
In the README file of our submitted code repo, you will find all the necessary bash scripts to reproduce main tables and figures from our experiments. The hyperparameter details are all in our code.

\section{Time Complexity and Convergence Discussions}

\subsection{Time Complexity Discussions}

We employ a smoothing proximal gradient method that is originally developed for structured-sparsity-inducing penalties. By using the efficient method, the convergence rate of the algorithm is $O(\frac{1}{\epsilon})$

\textbf{For sGLP-IB:} the time complexity per iteration of the smoothing proximal gradient for sGLP-IB is $O(in_i^2out_i+in_i|E|)$. Thus, the overall time complexity for sGLP-IB is  $O(\frac{1}{\epsilon}\times(in_i^2out_i+in_i|E|))$.


\textbf{For sTLP-IB:} since a tree associated with $out_i$ responses can have at most $2out_i-1$ nodes, which constrains the $|G_v|$. It is computationally efficient and spatially economical to run sTLP-IB. The time complexity per iteration of the smoothing proximal gradient is $O(in_i^2out_i+in_i\sum{v\in V}|G_v|)$. Given the desired accuracy $\epsilon$, the overall complexity for our method is $O(\frac{1}{\epsilon}\times(in_i^2out_i+in_i\sum{v\in V}|G_v|))$.  

\subsection{Convergence Discussions}
We rewrite $bs\times bs$ as $N$, $\bm{X}_i^g$ as $\bm{X}$,  $in_i$ as $J$, $out_i$ as $K$, $\bm{X}_{i+1}^g$ as $\bm{Y}$. Then we will get the original form of Lasso as the following: 

\begin{equation}
\label{equa:standardlasso2}
\bm{\beta} = \min  \limits_{\bm{\beta}}\frac{1}{2}||\bm{Y} - \bm{X}\bm{\beta}||_F^2 + \Phi(\bm{\beta})
\end{equation}

The edge set $E$ for sGLP-IB is generated similarly to graph-
structured lasso; the $G_v$ in sTLP-IB is also generated following tree-guided lasso. Thus, $\Phi(\bm{\beta})$ follows the original forms of the previous two lasso variants.  Based on these, our methods also follow the proximal gradient descent framework~\cite{chen2012smoothing,Yuan2024SmoothingPG} and are applied to optimization problems of the same format. The format is as follows:

\begin{equation}
\label{equa:format}
 \min  \limits_{x}f(x)+g(x)
\end{equation}

Where $f(x)$ ($\frac{1}{2}||\bm{Y} - \bm{X}\bm{\beta}||_F^2$) is a smooth and differentiable function, while $g(x)$ ($ \Phi(\bm{\beta})$) is non-smooth but has a known proximity operator~\cite{Chen2010GraphStructuredMR,Kim2009TreeGuidedGL}. For $\epsilon$ (the differences between the yielded $\bm{\beta}$ and the optimal $\bm{\beta}^{*}$), the proximal gradient optimization method could guarantee convergence within polynomial time complexity for the optimization problems following the format like \eqref{equa:format}. 

Thus, our methods, sGLP-IB and sTLP-IB, possess the property of convergence



\begin{figure}[!th]
  \centering
  \includegraphics[width=0.6\linewidth]{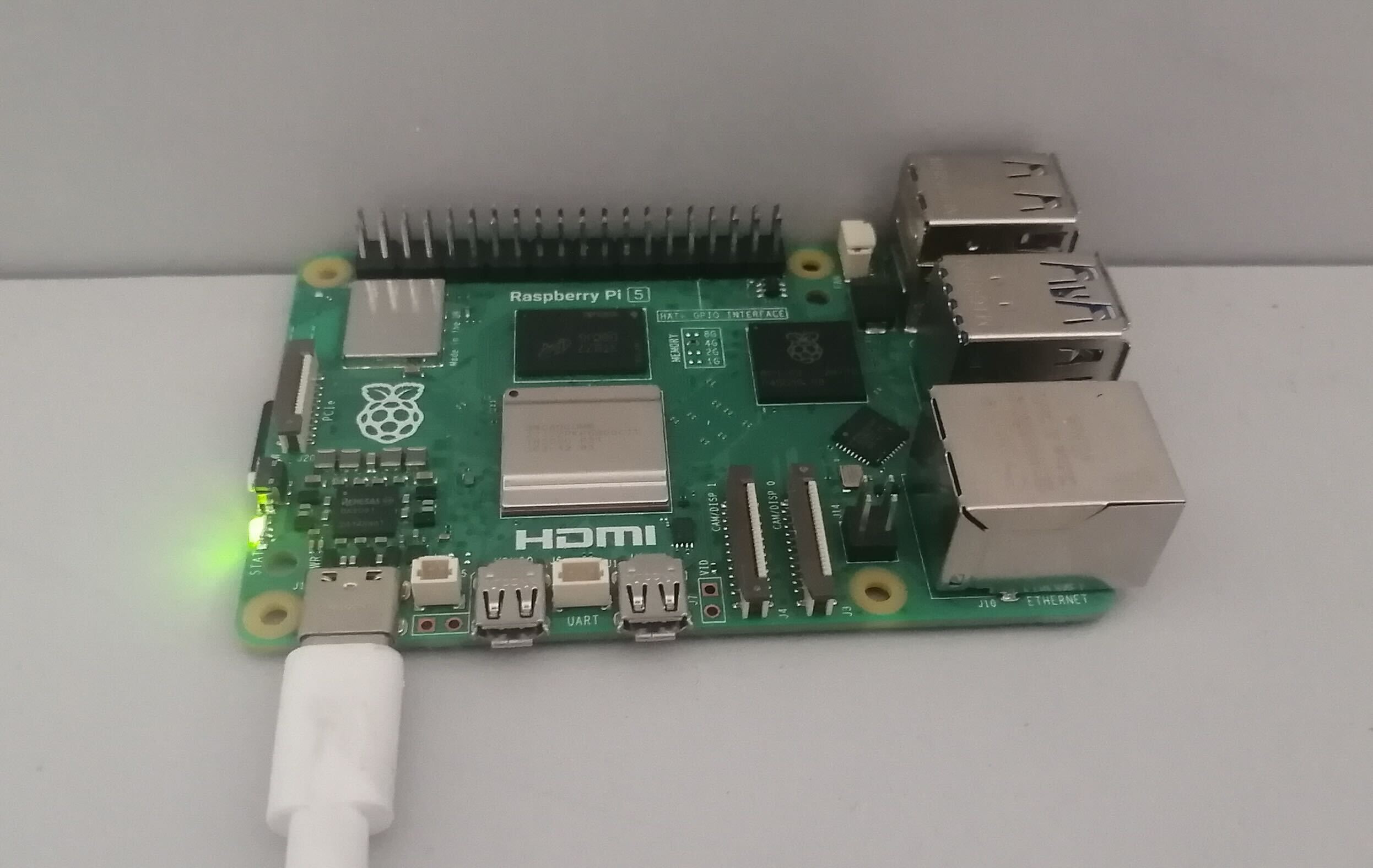}
  \caption{Our example experimental prototype for resource-constrained devices.}
  \label{fig:device}
\end{figure}

\section{Experiments}

\subsection{Adaption for Pruning ResNet and GoogleNet}
As architectures of ResNet and GoogleNet differ from conventional CNNs, we make adaptations to the pruning process accordingly.

\textbf{For ResNet:} due to ResNet's architecture featuring shortcuts, when pruning ResNet variants, we initially ignore the shortcuts. We perform pruning layer by layer, and then prune this shortcut layer corresponding to the pruned channels of its related layers inside the residual block when we approach the end of this residual block. This pruning approach ensures the integrity and functionality of the shortcut connections while effectively reducing the overall network's complexity.

\textbf{For GoogleNet:} we perform pruning on each branch of Inception individually using sGLP-IB and sTLP-IB.

\subsection{Dataset Information}
\textbf{Datasets: }We used three widely adopted datasets to evaluate our proposed methods against baselines: CIFAR10~\cite{Krizhevsky2009LearningML}, CIFAR100~\cite{Krizhevsky2009LearningML} and ImageNet~\cite{deng2009imagenet}. 

\begin{itemize}
    \item CIFAR-10 consists of 60,000 color images with $32 \times 32$ pixels, evenly distributed across 10 classes.
    \item CIFAR-100 dataset consists of 60,000 $32 \times 32$ color images, evenly distributed across 100 classes.
    \item ImageNet dataset contains 1.2 million $224\times224$ color images across 1,000 classes, with each class having around 1,000 images.
\end{itemize}

\begin{table}[!t]
\centering
\caption{Pruning results of VGGNet-16 on CIFAR-100 dataset.}
\resizebox{\linewidth}{!}{
\begin{tabular}{lcccc}
\toprule
\textbf{Method}  &\textbf{Top-1(\%)}    &\textbf{FLOPs}$\downarrow$ &\textbf{Params} $\downarrow$  &  $\Delta$ \textbf{Acc(\%)}   \\
\midrule
\texttt{VGGNet-16}&\texttt{73.80} & \texttt{0(314.57M)}  &\texttt{0(14.98M)}  &\texttt{0}      \\
CPMC	 &73.01	&48\%	&-	&-0.79\\
PGMPF	 &73.45	&48\%	&-	&-0.35\\
CPGMI	 &73.53	&37\%	&-	&-0.27\\
PGMPF-SFP&73.66	&48\%	&-	&-0.14\\
APIB	 &73.89	&48\%	&57	&+0.09\\
\textbf{sGLP-IB}	&\textbf{73.96}&\textbf{55\%}&\textbf{71\%}	&\textbf{+0.16}\\
\textbf{sTLP-IB}	&\textbf{73.99}&\textbf{55\%}&\textbf{71\%}	&\textbf{+0.19}\\
\bottomrule
\end{tabular}
}
\label{table:cifar100}
\end{table}

\subsection{Experiments with CIFAR-100 Dataset}
\label{sec:cifar100}
We compare sGLP-IB and sTLP-IB with current methods in Table~\ref{table:cifar100} on the CIFAR-100 dataset using VGGNet-16. Our methods, sGLP-IB and sTLP-IB, achieve the highest accuracy with the largest pruning ratio (55\% for FLOPs and 71\% for parameters), reaching 73.96\% and 73.99\%, respectively, which are 0.16\% and 0.19\% higher than the base model. This comparison with these state-of-the-art methods highlights the effectiveness of our approach, considering Top-1 accuracy, FLOPs pruned ratio, and parameter reduction ratio metrics. Combined with results from experiments on CIFAR-10 and ImageNet, it demonstrates that our methods can be extended to various real-world datasets. 

\begin{table}[!h]
\centering
\resizebox{\linewidth}{!}{
\begin{tabular}{lcccccc}
\toprule
\textbf{Ratio}  &40  & 50 & 60 & 70 & 80\\
\midrule
sGLP-IB & 94.12(45\%) & 94.09(54\%)& 94.04(84\%) & 93.74(93\%) & 91.00(94\%)\\
sTLP-IB & 94.12(47\%) & 94.11(55\%)& 94.09(84\%) & 93.80(93\%) & 91.75(94\%)\\
\bottomrule
\end{tabular}
}
\caption{Results of our methods when varying the FLOPs pruning ratios using VGGNet-16 on CIFAR-10 (The parameter ratio of pruned FLOPs is indicated in parentheses.).}
\label{table:ratioflop}
\end{table}

\subsection{Experiments with Pruning Ratios for FLOPs}
In Table~\ref{table:ratioflop}, before the pruning ratio exceeds 50\%, we observe that the parameter pruned ratio is also below 60\%, and the performance of sGLP-IB and sTLP-IB is nearly identical. It is noteworthy, however, that sTLP-IB prunes more parameters, indicating that it retains fewer parameters while maintaining the same FLOPs. This suggests that sTLP-IB is more effective at focusing on important filters due to its superior ability to model structured class-wise information. When the pruning ratio exceeds 70\%, the accuracy of both methods begins to drop. At a ratio of 80\%, the accuracy of our methods drops by 2-3\% due to excessive parameter reduction (94\%), though sTLP-IB still remains more robust than sGLP-IB.

\begin{table}[]
\caption{Results of our methods on a Raspberry Pi-4B device with four models on CIFAR-10. The values in parentheses indicate the percentage reduction in latency.}
\resizebox{0.95\linewidth}{!}{
\begin{tabular}{lcccc}
\toprule
Latency(s)&VGGNet-16&ResNet-56&ResNet-110&GoogleNet \\
\midrule
Original&26.7 &4.6 &7.7 &11.2\\
sGLP-IB &7.2(73\%) &2.2(52\%) &3.2(58\%) &3.8(66\%) \\
sTLP-IB &7.2(73\%) &2.2(52\%) &3.3(57\%) & 3.8(66\%)  \\
\bottomrule
\end{tabular}
}
\label{table:device}
\end{table}

\subsection{Implementation on Resource-constrained Devices}
In addition to accuracy, memory size, and FLOPs, we also evaluate the latency we reduced using our method on resource-constrained devices. We use a Raspberry Pi 4B device. The results are presented in Table~\ref{table:device}. We select the highest-accuracy models of our sGLP-IB and sTLP-IB methods for each of the four architectures listed in Table 1 of main paper. The results show that our method significantly reduces latency up to 73\% while enabling the deployment of complex neural models on the resource-constrained devices. 

\subsection{Resource-constrained Devices}

Our example experimental prototype Raspberry Pi 4B for resource-constrained devices is shown in Figure~\ref{fig:device}.

\end{document}